\documentclass[journal]{IEEEtran}

\usepackage{amsmath,amsfonts}
\usepackage{algorithmic}
\usepackage{algorithm}
\usepackage{array}
\usepackage[caption=false,font=normalsize,labelfont=sf,textfont=sf]{subfig}
\usepackage{textcomp}
\usepackage{stfloats}
\usepackage{url}
\usepackage{verbatim}
\usepackage{graphicx}
\usepackage{cite}
\usepackage{bbding}
\usepackage{float}
\usepackage{tabularx}
\usepackage{multirow}
\usepackage{xcolor}
\usepackage{colortbl}
\usepackage{booktabs}
\usepackage{bm}
\usepackage{cuted}
\usepackage{capt-of}
\usepackage[hidelinks]{hyperref}
\makeatletter
\providecommand{\@makespecialcolbox}{}
\makeatother
\definecolor{First}{rgb}{0.95, 0.62, 0.61}
\definecolor{Second}{rgb}{0.97,0.81,0.63}
\definecolor{Third}{rgb}{1.0, 0.97, 0.70}

\begin{document}

\title{PressMimic: Pressure-Guided Motion Capture and Control for Humanoid Robot Imitation}

\author{
Yi Lu$^{*1}$,~Shenghao Ren$^{*1}$,~Tianyu Xiong$^{1}$,~Zhaoxiang Li$^{1}$,~Jiaqi Li$^{1}$,\\
~He Zhang$^{3}$,~Tao Yu$^{3}$,~Qiu Shen$^{\dagger1,2}$,~Xun Cao$^{1,2}$
\thanks{$^*$Equal contribution. $^\dagger$Corresponding author. E-mail: shenqiu@nju.edu.cn}
\thanks{$^1$School of Electronic Science and Engineering, Nanjing University, Nanjing 210023, China.}
\thanks{$^2$Key Laboratory of Optoelectronic Devices and Systems with Extreme Performances of MOE, Nanjing University, Nanjing 210023, China.}
\thanks{$^3$BNRist, Tsinghua University, Beijing 100084, China.}
\thanks{Manuscript received May 24, 2026.}
}

\markboth{Journal of \LaTeX\ Class Files,~Vol.~14, No.~8, August~2021}%
{Shell \MakeLowercase{\textit{et al.}}: A Sample Article Using IEEEtran.cls for IEEE Journals}

\IEEEpubid{0000--0000/00\$00.00~\copyright~2021 IEEE}


\maketitle

\begin{strip}
\vspace{-80pt}
\centering
\includegraphics[width=\textwidth]{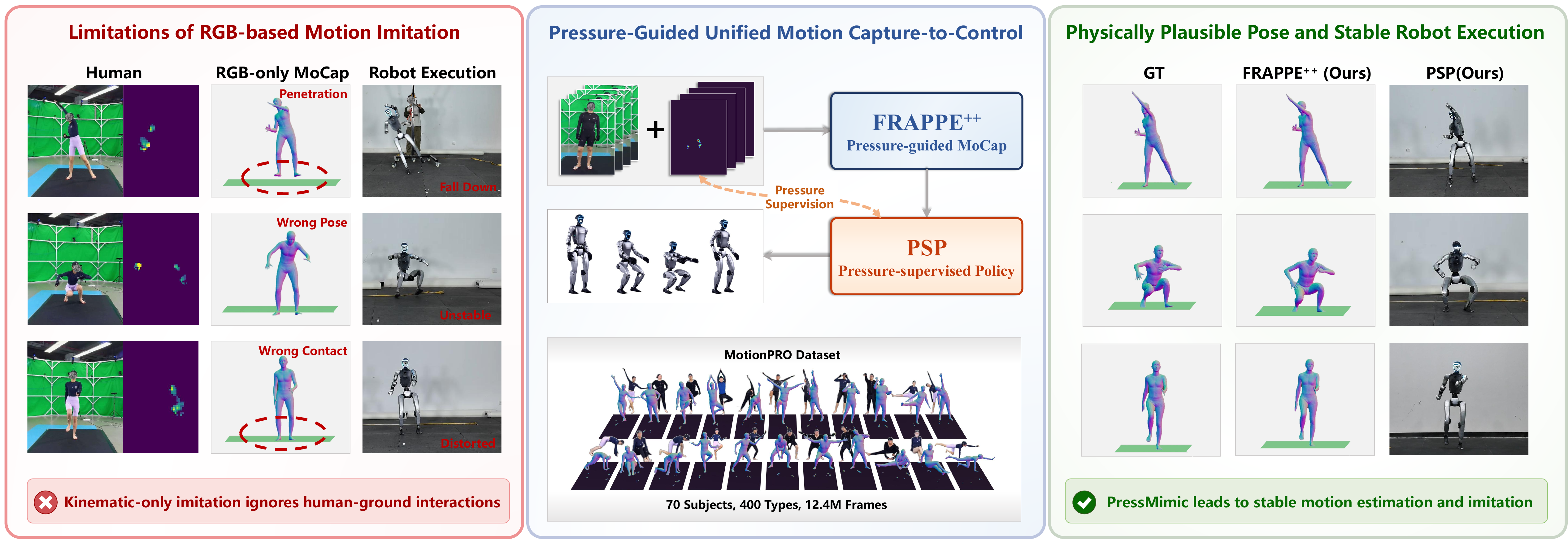}
\captionof{figure}{We propose PressMimic, a unified framework that integrates pressure into both motion capture and motion control for humanoid motion imitation. By fusing RGB with pressure signals in the motion capture module $\text{FRAPPE}^{++}$ and supervising robot training via the pressure-guided policy PSP, PressMimic achieves physically grounded pose estimation and stable robot execution.}
\label{fig:teaser}
\end{strip}

\begin{abstract}
Humanoid motion imitation requires not only accurate perception of human kinematics but also faithful reproduction of physical interactions with the environment. However, existing pipelines rely primarily on vision-based motion capture and kinematic imitation, largely ignoring contact dynamics, leading to artifacts such as foot sliding, floor penetration, and unstable behaviors.
In this work, we revisit humanoid motion imitation from the perspective of physical grounding and leverage pressure as a unified modality across perception and control. We present PressMimic, a framework that integrates pressure into the full pipeline from motion capture to humanoid control. In the perception stage, we introduce $\text{FRAPPE}^{++}$, a multimodal model that fuses RGB and pressure to jointly estimate 3D pose and global motion, where pressure provides explicit contact and support constraints to resolve ambiguity in vision-based estimation. In the control stage, we propose a pressure-supervised policy (PSP) that incorporates pressure-derived signals into reinforcement learning, enabling physically consistent contact patterns during execution.
We further construct MotionPRO, a large-scale dataset with synchronized RGB, pressure, and motion capture data. Experiments show that pressure improves motion estimation accuracy, trajectory consistency, and execution stability. These results demonstrate that pressure serves as an effective physical grounding signal, bridging perception and control for physically consistent humanoid motion imitation. \textit{Project Page: \url{https://yeelou.github.io/PressMimic/}}
\end{abstract}

\begin{IEEEkeywords}
Human Motion Capture, Dataset, Pressure Sensing, Multimodal Fusion, Humanoid Robot Control, Reinforcement Learning, Embodied Intelligence
\end{IEEEkeywords}
\IEEEpubidadjcol

\section{Introduction}
\IEEEPARstart{E}nabling humanoid robots to faithfully replicate human motion has long been a central challenge at the intersection of computer vision and robotics. Achieving this goal demands not only accurate perception of human body kinematics, but also robust translation of such perception into physically plausible robot control. Despite remarkable progress in both human motion capture~\cite{kanazawa2018end, tripathi20233d, kolotouros2019learning, li2022cliff,fang2021reconstructing,li2021hybrik,goel2023humans,huang2018deep,yi2022physical,du2023avatars} and humanoid motion control~\cite{gu2025humanoid,xie2020review,DBLP:journals/csur/Cao26}, existing approaches remain agnostic to the physical interactions between the body and its environment, leading to failure modes including temporal jitter, floor penetration, and foot-sliding artifacts. This perceptual deficiency creates a critical bottleneck: robots driven by kinematically-estimated references frequently fail to reproduce the physical grounding of human motion, resulting in unstable locomotion, degraded performance in scenarios with frequent contact changes, and, in extreme cases, hardware damage.

Pressure naturally encodes the foot-ground interactions and contact dynamics that kinematic signals fail to capture. Previous work~\cite{DBLP:conf/cvpr/RenLHZZYSC25} demonstrated that pressure substantially improves MoCap quality by providing explicit grounding constraints on 3D pose and trajectory estimation. Beyond perception, pressure also offers a direct supervisory signal for robot motion control, constraining contact timing, ground reaction forces, and weight distribution that existing imitation pipelines overlook.

In this work, we present \textbf{PressMimic}, a unified framework that leverages pressure as a core modality across the full pipeline of humanoid motion imitation from human motion capture to robot motion control. To the best of our knowledge, PressMimic is the first system to systematically explore and exploit pressure signals in both the perception and control stages of humanoid motion imitation. 

The proposed framework operates in two stages. In the first stage, we introduce a multimodal human motion estimation network \textbf{$\text{FRAPPE}^{++}$} that fuses RGB video with pressure maps. Both modalities are non-invasive and require no wearable instrumentation, yet capture complementary aspects of human motion: RGB encodes visual appearance and geometric structure, while pressure provides explicit contact states and ground reaction dynamics imperceptible to cameras alone. To exploit this complementarity, we extract compact pressure representations and model spatiotemporal dependencies across modalities, fusing them into a unified motion representation. We further impose orthographic similarity constraints along the camera axis and whole-body contact constraints along the vertical axis to combine precise lower-body contact and global translation cues from pressure with accurate full-body pose from RGB. In the second stage, we introduce a Pressure-Supervised motion control Policy (\textbf{PSP}), where plantar pressure maps are modeled as distributional offsets characterizing foot contact patterns and incorporated as an auxiliary reward term during reinforcement learning, explicitly encouraging the robot to reproduce the ground reaction dynamics of the human demonstrator. 

To support both stages, we introduce \textbf{MotionPRO}, a large-scale multimodal dataset capturing pressure, RGB video, and optical motion capture signals from 70 volunteers performing 400 motion types, encompassing 12.4M pose frames in total. Extensive experiments on MotionPRO demonstrate that our multimodal fusion approach outperforms state-of-the-art RGB-based pose and trajectory estimation methods, maintaining physically plausible estimates even under extreme occlusions and severe vertical trajectory drift. For humanoid motion imitation, pressure supervision consistently improves task success rates and locomotion stability while preserving strong motion similarity to the reference demonstrations.

The main contributions of this work are summarized as follows:
\begin{itemize}
\item We present \textbf{PressMimic}, the first unified framework to systematically exploit pressure across both the perception and control stages of humanoid motion imitation.
\item We propose a multimodal motion estimation network \textbf{$\text{FRAPPE}^{++}$} fusing RGB and pressure, achieving robust pose and trajectory estimation under occlusion and trajectory drift.
\item We introduce a pressure-supervised control policy called \textbf{PSP} that incorporates pressure-derived distributional offsets as an auxiliary reward, improving task success rates and locomotion stability in humanoid motion imitation.
\item We construct \textbf{MotionPRO}, a large-scale human motion dataset with pressure, RGB, and optical sensors. Extensive experiments demonstrate that our methods can achieve significant performance improvements in both motion estimation and motion imitation tasks.
\end{itemize}

\section{Related Work}

\subsection{Humanoid Robot Motion Imitation}

Humanoid robot motion imitation aims to enable robots to reproduce human motion in a physically plausible and stable manner. Early approaches~\cite{koenemann2014real,abi2018humanoid,elobaid2019telexistence,ayusawa2017motion,otani2017adaptive,penco2018robust,darvish2019whole,dafarra2024icub3} relied on motion retargeting to directly map human joint configurations to robot joint angles, followed by trajectory optimization or model predictive control to generate feasible robot motions. While effective for simple periodic motions such as walking, these methods struggle to generalize to complex or contact-rich behaviors due to their reliance on accurate dynamic models and hand-crafted controllers.

Reinforcement learning (RL) has emerged as the dominant paradigm for humanoid motion imitation, enabling controllers to be learned directly from interaction with physics simulators. DeepMimic~\cite{peng2018deepmimic} pioneered the use of reference-state initialization and motion imitation rewards to train physics-based controllers capable of reproducing a wide range of human motions. Subsequent works extended this framework to handle more diverse motion repertoires~\cite{luo2023perpetual,luo2024universal}, multi-skill transitions~\cite{tessler2024maskedmimic,luo2024omnigrasp}, and real-world deployment on physical robots~\cite{yang2025omniretarget,zhang2025track,luo2025sonic,he2024omnih2o,xie2025kungfubot,he2025asap,li2025bfm,ze2025twist2}. PHC~\cite{luo2023perpetual} introduced a perpetual humanoid controller that tracks arbitrary motion sequences through curriculum-based training. More recently, BeyondMimic~\cite{liao2025beyondmimic} and SONIC~\cite{luo2025sonic} have advanced whole-body motion imitation on full-sized humanoid platforms, demonstrating stable policy deployment on real hardware.

Despite these advances, existing methods construct their imitation rewards purely from kinematic references, including joint positions, orientations, and end-effector states, without accounting for the dynamic foot-ground interactions that underpin stable locomotion. This limitation frequently manifests as foot-sliding, floor penetration, and unstable gait patterns, particularly in contact-rich scenarios. In contrast, PressMimic incorporates plantar pressure as an explicit contact supervision signal, encouraging the robot to reproduce not only the kinematic trajectory but also the underlying contact dynamics of the human demonstrator.

\subsection{Non-intrusive Human Motion Capture}

Robotic imitation learning relies heavily on human motion capture to acquire high-quality demonstrations of human behaviors. Traditional MoCap systems, such as marker-based optical setups~\cite{DBLP:journals/sensors/MerriauxDBVS17}, as well as approaches based on wearable IMUs~\cite{faisal2019review,yi2022physical}, VR tracking devices~\cite{DBLP:conf/siggrapha/WinklerWY22}, can provide precise measurements but typically require markers or sensors to be attached to the human body. This not only increases system complexity and cost but also interferes with natural human motion. To address these limitations, non-intrusive human pose estimation has emerged as a promising alternative. It aims to recover 3D human pose and motion trajectories from external observations without requiring any markers or on-body sensors. Compared with conventional MoCap systems, non-intrusive approaches offer advantages such as flexible deployment, lower cost, and minimal interference with users, making them increasingly attractive for robotic applications~\cite{fu2024humanplus,he2024omnih2o,DBLP:journals/corr/abs-2412-13196,he2025asap}.

Existing non-intrusive motion capture methods mainly rely on visual perception based on RGB inputs, early approaches typically regress 3D joint positions or parametric human body models from single RGB images ~\cite{kanazawa2018end, kolotouros2019learning, li2021hybrik, tripathi20233d, li2022cliff, fang2021reconstructing, goel2023humans}. With the advancement of deep learning, an increasing number of methods utilize the perspectives of temporal information~\cite{kocabas2020vibe, choi2021beyond, kanazawa2019learning, sun2019human, arnab2019exploiting}, prior knowledge of human body~\cite{bogo2016keep, kanazawa2018end, pavlakos2019expressive, zanfir2020weakly}, and precise camera model~\cite{li2022cliff, kissos2020beyond, kocabas2021spec, wang2023zolly} to obtain accurate human body pose. 
However, purely vision-based methods remain inherently ill-posed due to depth ambiguity and occlusion, which often lead to global scale inconsistency and physical artifacts such as foot skating or floor penetration. Moreover, they struggle to infer human--environment interactions, especially contact states, which are critical for physically plausible motion estimation~\cite{yuan2022glamr, ye2023decoupling, wang2024tram, sun2023trace, shin2024wham, shen2024gvhmr}. 

Ground pressure sensors can directly capture the contact states and force distributions between the human body and the environment, offering unique advantages in modeling foot contact, support patterns, and center-of-mass dynamics. Early works reconstruct human pose from single-frame pressure data, mainly focusing on lying scenarios~\cite{clever20183d, clever2020bodies, yin2022multimodal, tandon2024bodymap, liu2022simultaneously, clever2022bodypressure, wu2024seeing}. Moreover, auxiliary information derived from pressure data, such as the center of pressure (CoP), is used to constrain pose estimation and improve physical plausibility~\cite{tripathi20233d, zhang2024mmvp}. Despite these advances, existing pressure-based methods still suffer from several limitations. On the one hand, CoP-based methods only work well for slow and steady movements. They fail in fast actions because they don't account for the inertia that comes with rapid motion.~\cite{han2023groundlink, mourot2022underpressure}. On the other hand, most approaches focus on foot contact and lack modeling of contact interactions involving other body parts~\cite{scott2020image, luo2021intelligent}. Therefore, effectively leveraging the physical priors provided by pressure signals to develop more robust and physically plausible non-intrusive human motion capture methods remains a critical challenge.

\section{Methods}
\subsection{Pressure-Guided Humanoid Motion Imitation}

\begin{figure*}[htb]
	\centering
	\includegraphics[width=\linewidth]{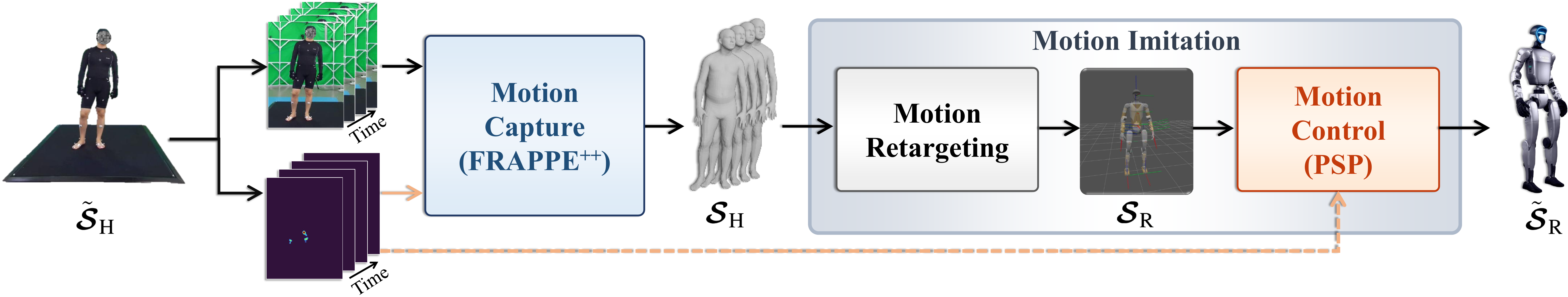}
	\caption{\textbf{Overview of PressMimic.} Given synchronized RGB video and pressure sequences, PressMimic estimates human body motion, retargets it to the humanoid robot, and executes it via a learned control policy. Pressure signals enhance motion estimation accuracy and provide auxiliary contact supervision for the control policy, as indicated by the orange dashed arrow.}
	\label{fig:pipeline}
\end{figure*}

As illustrated in Fig.~\ref{fig:pipeline}, the proposed PressMimic framework takes synchronized RGB video and pressure sequences as input and produces humanoid robot motion as output, operating through three successive modules: motion estimation, motion retargeting, and motion control. Pressure signals serve as a core modality throughout the pipeline, enhancing both the accuracy of human motion estimation and the physical plausibility of robot motion control.

The humanoid motion imitation task can be formulated as an optimization problem over human motion parameters and robot joint configurations:
\begin{equation}
    \label{eq:problem_reform}
    \begin{aligned}
    \min \limits_{\bm{\theta}, \bm{T}, \bm{A}}
        \quad \mathcal{L}_{\text{H}}(&\bm{\mathcal{S}}_{\text{H}}(\bm{\theta},\bm{T}), \tilde{\bm{\mathcal{S}}}_{\text{H}}) + \mathcal{L}_{\text{R}}(\bm{\mathcal{S}}_{\text{R}}(\bm{A}), \tilde{\bm{\mathcal{S}}}_{\text{R}}(\tilde{\bm{A}})) \\
        \text{s.t.}\quad\quad 
        &\bm{\mathcal{S}}_{\text{R}} = \mathcal{M}(\bm{\mathcal{S}}_{\text{H}}) \\
        &\tilde{\bm{\mathcal{S}}}_{\text{R}} \subseteq \bm{C}_{\text{R}}
    \end{aligned}
\end{equation}
where $\bm{\mathcal{S}}_{\text{H}}(\bm{\theta}, \bm{T})$ denotes the full state representation of the human body model parameterized by pose $\bm{\theta}$ and global translation $\bm{T}$. $\tilde{\bm{\mathcal{S}}}_{\text{H}}$ denotes the corresponding ground-truth human state. $\mathcal{L}_{\text{H}}$ is a composite loss function measuring the discrepancy between estimated and ground-truth human states across multiple terms including pose, joint positions and foot contact states. $\bm{\mathcal{S}}_{\text{R}}(\bm{A})$ denotes the reference robot state with action $\bm{A}$, and $\tilde{\bm{\mathcal{S}}}_{\text{R}}(\tilde{\bm{A}})$ denotes the actually executed robot state with action $\tilde{\bm{A}}$. $\mathcal{L}_{\text{R}}$ is a composite distance function over the robot state space, including body positions, orientations, velocities, and angular velocities. $\mathcal{M}(\cdot)$ denotes the mapping from the human body model to the humanoid robot model, and $\bm{C}_{\text{R}}$ denotes the physical feasibility constraints of the robot, including joint angle limits, velocity bounds, and stability conditions.

The first term of the objective minimizes the discrepancy between the estimated human body joints and their ground-truth counterparts, which is addressed in the motion estimation stage (Sec.~\ref{sec:estimation}). The second term minimizes the gap between the reference robot joints, retargeted from the estimated human motion via $\mathcal{M}(\cdot)$, and the actually executed robot joints, which is addressed in the motion control stage (Sec.~\ref{sec:control}). The constraint $\tilde{\bm{S}}_{\text{R}}(\tilde{\bm{A}}) \subseteq \bm{C}_{\text{R}}$ ensures that the executed robot motion remains within the physical feasibility bounds of the robot throughout imitation. Pressure signals are incorporated into both stages as indicated by the orange dashed arrow in Fig.~\ref{fig:pipeline}, they provide grounding constraints during motion estimation and serve as auxiliary supervision during motion control.

\subsection{Human Motion Estimation by Fusing Pressure and RGB}
\label{sec:estimation}

\begin{figure*}[htb]
	\centering
	\includegraphics[width=0.95\linewidth]{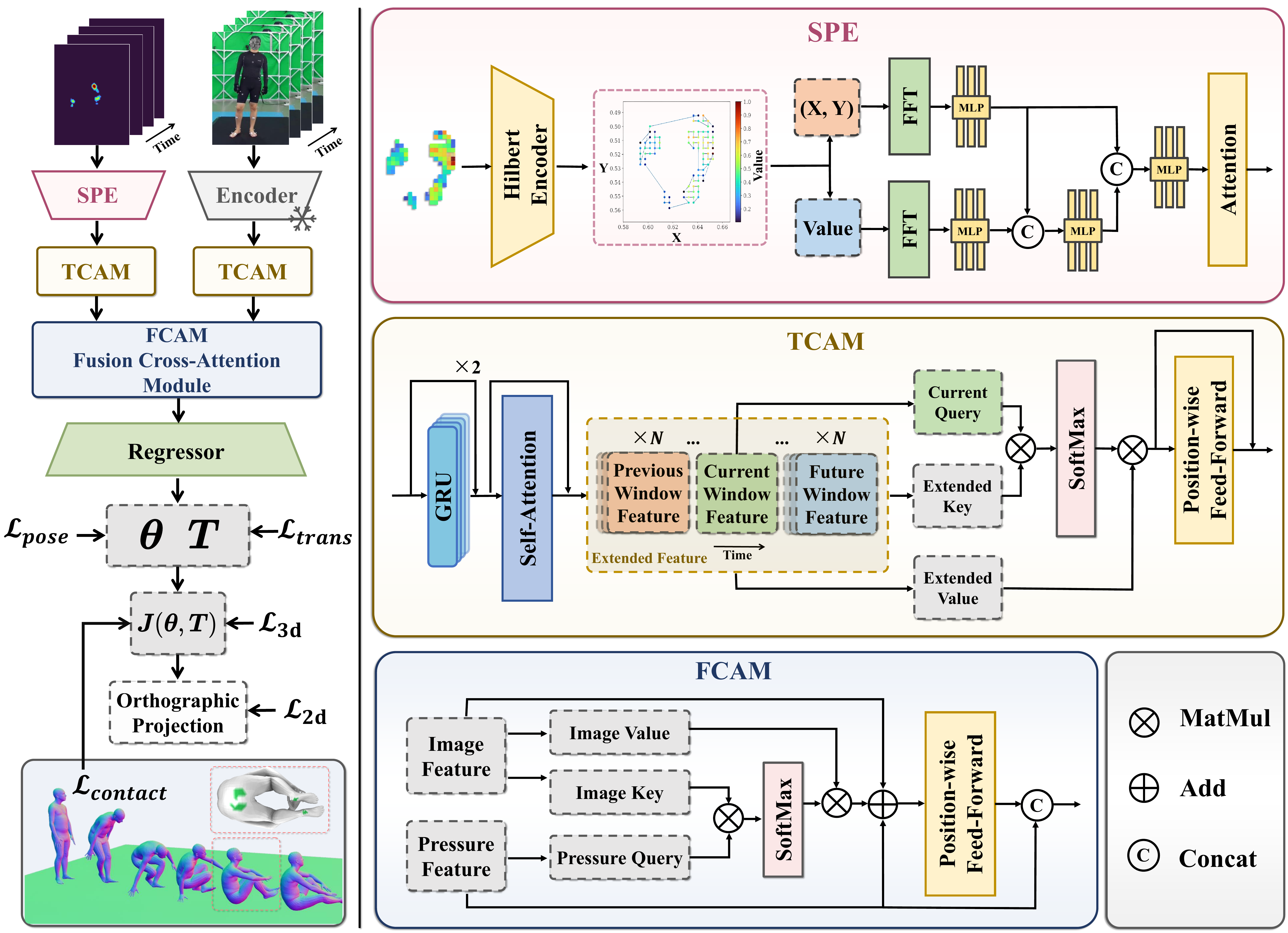}
	\caption{\textbf{The framework of $\text{FRAPPE}^{++}$.} Pressure and RGB video are processed by the SPE and image encoder respectively, followed by TCAM to model spatiotemporal dependencies within each modality. FCAM then fuses the two modalities via cross-attention, where image features serve as key and value while pressure features serve as query. The fused representation is fed into a regressor to estimate SMPL parameters.}
	\label{pose_estimation}
\end{figure*}

In the first stage, we present a pressure-guided motion estimation framework $\text{FRAPPE}^{++}$ that integrates pressure representation learning, temporal context modeling, and cross-modal feature fusion to achieve robust estimation of human 3D pose and global motion trajectories, as illustrated in Fig.~\ref{pose_estimation}.

\subsubsection{Pressure-Guided Motion Estimation Pipeline}
Given a synchronized RGB video sequence and a corresponding pressure distribution sequence, our goal is to recover the human 3D pose parameters $\bm{\theta}$ and the global translation trajectory $\bm{T}$ from multimodal observations.
First, the input RGB images are processed by a visual encoder to extract frame-wise visual features that represent human appearance and pose structure. Meanwhile, the two-dimensional pressure distribution maps are fed into the pressure encoder, referred to as \textbf{Sparse Pressure Encoder (SPE)}. This module converts the sparse pressure distribution into a structured one-dimensional sequence representation and extracts corresponding pressure features. The resulting pressure and visual feature sequences are then fed into the \textbf{Temporal Context Aggregation Module (TCAM)} for temporal context modeling, capturing temporal dependencies within the current window while incorporating contextual information from neighboring time windows. After temporal modeling, we introduce the \textbf{Fusion Cross-Attention Module (FCAM)} to enable cross-modal feature interaction. Finally, the temporally modeled features are fed into a regression network to predict the human pose parameters $\bm{\theta}$ and the global translation $\bm{T}$. The predicted parameters are further passed through a parametric human body model to compute the corresponding three-dimensional joint positions.

\subsubsection{Training Objective}
To achieve accurate pose reconstruction and physically consistent motion generation, we design a set of joint loss functions to simultaneously constrain the pose parameters, spatial joint positions, and global motion trajectories. The overall training objective is defined as
\begin{equation}
\begin{aligned}
 \mathcal{L}_{\text{H}}=
&\lambda_{\text{pose}}\mathcal{L}_{\text{pose}}+\lambda_{\text{3d}}\mathcal{L}_{\text{3d}} +\lambda_{\text{2d}}\mathcal{L}_{\text{2d}}  \\
&+\lambda_{\text{trans}}\mathcal{L}_{\text{trans}}+\lambda_{\text{contact}}\mathcal{L}_{\text{contact}}, 
\label{eq:frappe_loss}
\end{aligned}
\end{equation}
where $\lambda_{\mathrm{pose}}$, $\lambda_{\mathrm{3d}}$, $\lambda_{\mathrm{2d}}$, $\lambda_{\mathrm{trans}}$, and $\lambda_{\mathrm{contact}}$ are corresponding weights. The loss of pose parameters $\mathcal{L}_{\mathrm{pose}}$ is the mean squared error between the predicted and ground-truth pose parameters. The 3D joint loss, $\mathcal{L}_{\mathrm{3d}}$, is the mean squared error between the predicted and ground-truth joint positions, after pelvis alignment. Global translation loss $\mathcal{L}_{\mathrm{trans}}$ is the mean squared error between predicted and ground truth translation. The ground contact loss, $\mathcal{L}_{\mathrm{contact}}$, is the mean squared error between the predicted and the ground-truth global joints that are in contact with the ground.

\subsubsection{Sparse Pressure Encoder (SPE)}
In human pose estimation, ground pressure maps are highly sparse, making direct convolution over the full 2D map computationally inefficient and prone to missing fine-grained contact structures. To address this, we propose the Sparse Pressure Encoder (SPE), which converts the 2D pressure distribution into a compact 1D sequence while preserving spatial topology. 
Specifically, a Hilbert space-filling curve serializes the valid pressure points, maintaining local neighborhood relationships during the 2D-to-1D mapping. Each point is represented by its spatial coordinates and pressure value, encoded via Fourier sine-cosine embeddings to capture multi-scale spatial variations. The embedded features are projected by an MLP and refined through an attention mechanism that models relationships among all pressure points, yielding a fixed-length pressure feature vector for subsequent cross-modal fusion.

\subsubsection{Temporal Context Aggregation Module (TCAM)}
Accurate human motion estimation requires capturing temporal dependencies across long sequences. In practice, video sequences are often divided into fixed-length temporal windows for processing, which may introduce temporal discontinuities at window boundaries. 
To ensure the stability of motion estimation while managing computational overhead, TCAM adopts a modeling strategy that operates across two distinct scales. Within each window, we integrate Gated Recurrent Units with self-attention mechanisms to capture temporal dependencies that span short durations. Outside the current window, neighboring windows are compressed into compact representations through temporal average pooling. These representations can be introduced as augmented context via cross attention, effectively extending the field of reception for the model.

We divide the input sequence into a series of temporal windows and denote the current window as $\bm{x}_{t} = \{\bm{x}_{t}^{j}\}_{j=1}^{L}$, where $L$ is the number of frames per window, $\bm{x}_{t}^{j} \in \mathbb{R}^{D}$ is the feature vector of the $j$-th frame and $D$ is the feature dimension. 
To extend the temporal receptive field, we incorporate $N$ neighboring windows on each side of the current window. Each neighboring window $\bm{x}_{t+i}$ $(i \in [-N, N],\ i \neq 0)$ is compressed into a global representation $\bm{e}_{t+i} \in \mathbb{R}^{D}$ via temporal average pooling:
\begin{equation}
\bm{e}_{t+i} = \frac{1}{L} \sum_{j=1}^{L} \bm{x}_{t+i}^{j}.
\end{equation}
The compressed context representations are then concatenated with the current window to form an extended feature sequence:
\begin{equation} \bm{f}_t = [\bm{e}_{t-N}, \ldots, \bm{e}_{t-1}, \bm{x}_t, \bm{e}_{t+1}, \ldots, \bm{e}_{t+N}] . \end{equation}
Given $\bm{f}_t\in \mathbb{R}^{(L+2N) \times D}$, we apply a cross-attention mechanism where the current window $\bm{x}_t$ serves as the query source and the extended sequence $\bm{f}_t$ provides the keys and values. Specifically, the query $\bm{Q}_{\text{TCAM}}$, key $\bm{K}_{\text{TCAM}}$, and value $\bm{V}_{\text{TCAM}}$ are computed as:
\begin{equation}
\begin{aligned}
\bm{Q}_{\text{TCAM}} &= \bm{W}_{\text{q}} \bm{x}_t, \\
\bm{K}_{\text{TCAM}} &= \bm{W}_{\text{k}} \bm{f}_t, \\
\bm{V}_{\text{TCAM}} &= \bm{W}_{\text{v}} \bm{f}_t.
\end{aligned}
\end{equation}
where $\bm{W}_{\text{q}}, \bm{W}_{\text{k}}, \bm{W}_{\text{v}}$ are learnable projection matrices. The attention output is computed as:
\begin{align}
\bm{f}_{\text{TCAM}} & =
\mathrm{Attention}(\bm{Q}_{\text{TCAM}}, \bm{K}_{\text{TCAM}}, \bm{V}_{\text{TCAM}}) .
\end{align}
This allows each frame in $\bm{x}_t$ to attend to both fine-grained frame-level features within the current window and compressed global representations from neighboring windows, improving the temporal coherence of the predicted motion.

\subsubsection{Fusion Cross-Attention Module (FCAM)}
Visual information provides cues about human appearance and pose structure, whereas pressure signals capture contact states between the body and the ground. To exploit their complementary properties, FCAM performs cross-modal feature fusion.
Let $\bm{f}_{\text{rgb}}$ and $\bm{f}_{\text{p}}$ denote the output features of the visual encoder and the SPE module, respectively. Pressure features serve as the query source to actively probe the visual representation, while visual features provide the keys and values:
\begin{equation}
\begin{aligned}
\bm{Q}_{\text{FCAM}} &= \bm{W}_{\text{q}} \bm{f}_{\text{p}}, \\
\bm{K}_{\text{FCAM}} &= \bm{W}_{\text{k}} \bm{f}_{\text{rgb}}, \\
\bm{V}_{\text{FCAM}} &= \bm{W}_{\text{v}} \bm{f}_{\text{rgb}}.
\end{aligned}
\end{equation}
The fused representation is obtained by:
\begin{equation}
\bm{f}_{\text{FCAM}} = \mathrm{Attention}(\bm{Q}_{\text{FCAM}}, \bm{K}_{\text{FCAM}}, \bm{V}_{\text{FCAM}}).
\end{equation}
$\bm{f}_{\text{FCAM}}$ is subsequently passed through a position-wise feed-forward network with residual connections and concatenated with $\bm{f}_{\text{p}}$ to form the final motion representation. Through this design, pressure features guide visual attention toward human-ground contact regions, improving the accuracy and physical plausibility of motion estimation.

\begin{figure}[t]
	\centering
	\includegraphics[width=0.9\linewidth]{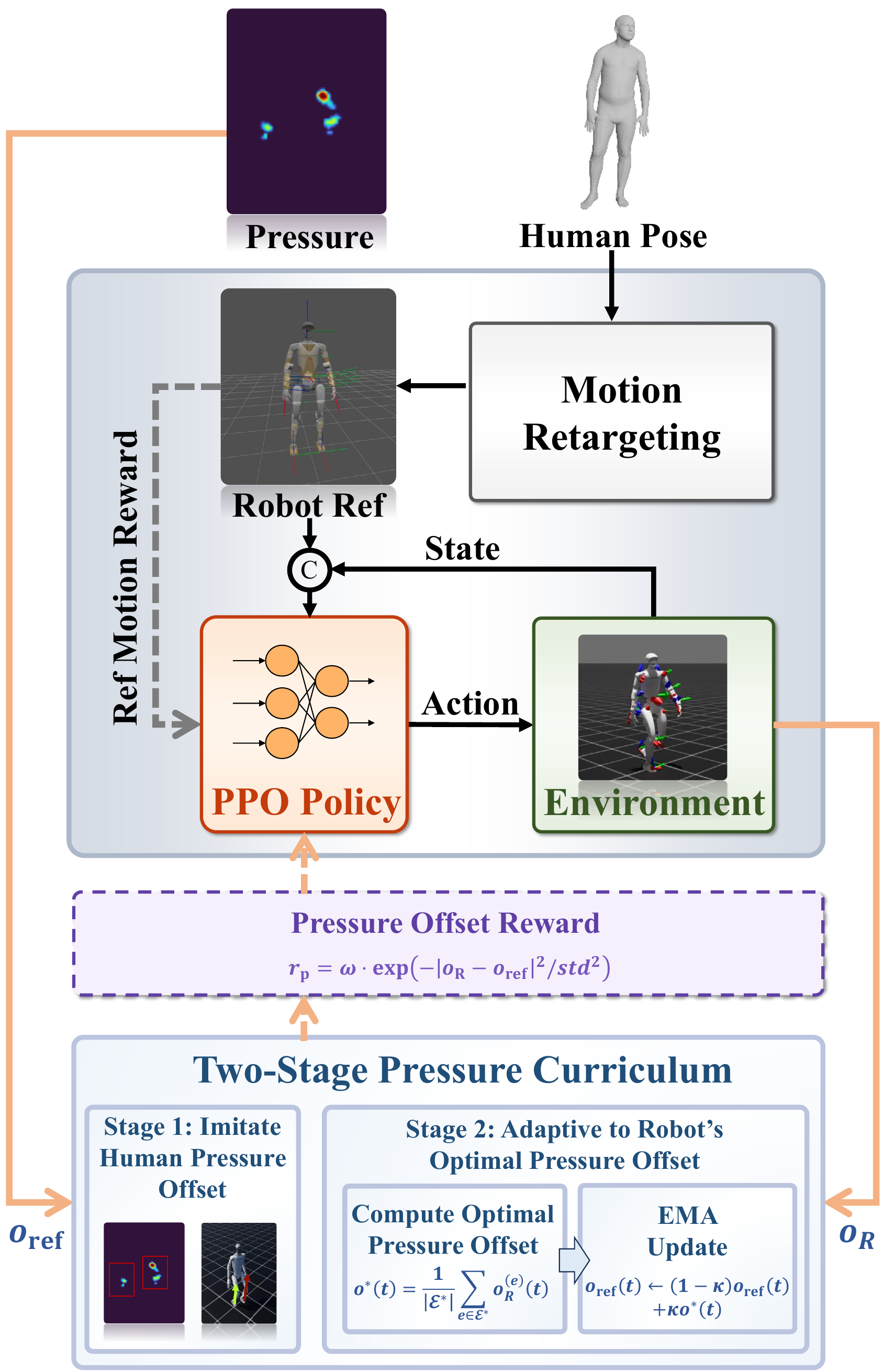}
	\caption{\textbf{Humanoid motion imitation under pressure-supervised policy (PSP).} Human pose and plantar pressure serve as dual inputs: the pose is retargeted to robot reference motion and concatenated with the current state as policy input, while pressure drives a two-stage curriculum that first imitates the human pressure offset and then adapts to the robot's own optimal contact distribution via EMA updates.}
	\label{fig:motion_control}
\end{figure}

\subsection{Humanoid Motion Control under Pressure Supervision}
\label{sec:control}
Building upon the pressure-guided motion estimation described above, we further leverage plantar pressure as an explicit supervisory signal to improve the physical consistency of humanoid motion control called PSP (Pressure-Supervised Policy), as illustrated in Fig.~\ref{fig:motion_control}.

\subsubsection{Pressure Modeling}

To incorporate plantar pressure into the control pipeline, we first extract per-foot pressure intensity from the raw pressure maps. Taking the left foot as an example, the total pressure of the left foot $\phi_{\text{lf}}$ is computed by summing the pressure values of all pixels belonging to the left foot region:
\begin{equation}
    \phi_{\text{lf}} = \sum_{i \in \text{lf}} \phi_{i},
\end{equation}
where $\phi_{i}$ denotes the pressure value of the $i$-th pixel. The right foot pressure $\phi_{\text{rf}}$ is computed analogously.

Building upon prior work~\cite{DBLP:conf/mm/LuRSC24}, we further define a pressure offset $o$ to characterize the bilateral weight distribution between the left and right feet:
\begin{equation}
\label{eq:op}
    o = \frac{\phi_{\text{rf}} + k \cdot \epsilon}{\phi_{\text{lf}} + \phi_{\text{rf}} + \epsilon},
\end{equation}
where $\epsilon > 0$ is a small constant for numerical stability. $k$ is a scalar that we set to $2$ in practice, chosen to place the airborne signature well outside the single-foot contact range $(0, 1)$ and far beyond any value reachable through normal load variation between the two feet. Because $\epsilon$ is negligibly small relative to any non-zero foot pressure, the numerical behaviour of $o$ is determined entirely by the foot pressure magnitudes in all contact cases, and no ambiguity arises at the boundary between contact states. This formulation provides an interpretable scalar representation of foot contact state. Specifically, when both feet are in contact ($\phi_{\text{lf}} \neq 0$ and $\phi_{\text{rf}} \neq 0$), $o$ lies within $(0, 1)$, reflecting the relative bilateral load distribution. When only the right foot is in contact ($\phi_{\text{lf}} = 0$ and $\phi_{\text{rf}} \neq 0$), $o$ approaches $1$. When only the left foot is in contact ($\phi_{\text{lf}} \neq 0$ and $\phi_{\text{rf}} = 0$), $o$ approaches $0$. When both feet are airborne ($\phi_{\text{lf}} = 0$ and $\phi_{\text{rf}} = 0$), $o = k = 2$, providing a distinctive out-of-contact signature that is unambiguously separable from all single-foot and double-foot contact cases.

\subsubsection{Policy Training with Pressure Supervision}

As illustrated in Fig.~\ref{fig:motion_control}, the motion control stage follows a reinforcement learning (RL) framework built upon BeyondMimic~\cite{liao2025beyondmimic}, trained with the Proximal Policy Optimization (PPO) algorithm~\cite{DBLP:journals/corr/SchulmanWDRK17}. The estimated human pose is first mapped to robot reference motion via motion retargeting module~\cite{DBLP:journals/corr/abs-2510-02252,zhao2026make}, which is concatenated with the current robot state as the policy input. The policy outputs joint-level actions that are executed in a physics simulator, and the resulting state is fed back to close the control loop.

The reward function comprises standard kinematic imitation terms adopted in prior works, including body joint position, joint angle, and end-effector rewards, supplemented by a pressure offset reward. To compute the pressure offset reward, we collect the ground reaction forces at the robot's feet from the simulator in real time, compute the robot's pressure offset $o_{\text{R}}$ following Eq.~\ref{eq:op}, and compare it against the reference pressure offset $o_{\text{ref}}$ derived from the human demonstrator's plantar pressure measurements. The pressure offset reward is defined as:
\begin{equation}
    r_{\text{p}} = \omega \cdot \exp\left(-\left| o_{\text{R}} - o_{\text{ref}} \right|^2 / std^2 \right),
\end{equation}
where $o_{\text{R}}$ and $o_{\text{ref}}$ denote the pressure offsets computed from the robot's ground reaction forces and the human demonstrator's plantar pressure measurements, respectively, following Eq.~\ref{eq:op}. $\omega$ is a weighting coefficient that scales the contribution of the pressure reward relative to other reward terms, and $std$ controls the sensitivity of the reward to deviations in pressure offset. This Gaussian-shaped reward provides dense and smooth supervision: it yields a high reward when the robot's bilateral weight distribution closely matches that of the human demonstrator, and decays gracefully as the discrepancy grows, avoiding abrupt reward discontinuities that could destabilize policy training.

To further improve training stability and adaptability, we introduce a two-stage pressure curriculum. In the first stage, the policy is trained with the fixed human reference pressure $o_{\text{ref}}$. Once the mean episode reward exceeds a preset threshold, the second stage is triggered, in which the reference pressure is dynamically updated to reflect the robot's own optimal contact distribution rather than the human demonstrator's. At each simulation step, the per-frame pressure offset $o_{\text{R}}$ is recorded across all parallel environments indexed by motion frame $t$. At the end of each episode, environments are ranked by total reward and the top $n$-th percentile is selected as the set of optimal environments $\mathcal{E}^*$, filtering out low-quality episodes that could corrupt the reference. The optimal pressure curve is obtained by averaging the per-frame pressure data across $\mathcal{E}^*$:
\begin{equation}
    o^*(t) = \frac{1}{|\mathcal{E}^*|} \sum_{e \in \mathcal{E}^*} 
    o_{\text{R}}^{(e)}(t),
    \label{eq:best_pressure}
\end{equation}
where frames not covered by any episode in $\mathcal{E}^*$ are excluded from the update. The reference pressure is then updated via exponential moving average (EMA):
\begin{equation}
    o_{\text{ref}}(t) \leftarrow (1 - \kappa)\, o_{\text{ref}}(t) 
    + \kappa\, o^*(t),
    \label{eq:ema}
\end{equation}
where $\kappa$ is the smoothing factor, ensuring the reference drifts gradually toward the robot's optimal behavior while avoiding training instability. After each update, the pressure history is cleared in preparation for the next episode. This curriculum transitions the supervision target from mimicking the human demonstrator to adapting to the robot's own stable contact dynamics, improving both imitation stability and task success rate.

\section{Experiments}
\subsection{Dataset}

\begin{figure}[t]
\centering
\includegraphics[width=\linewidth]{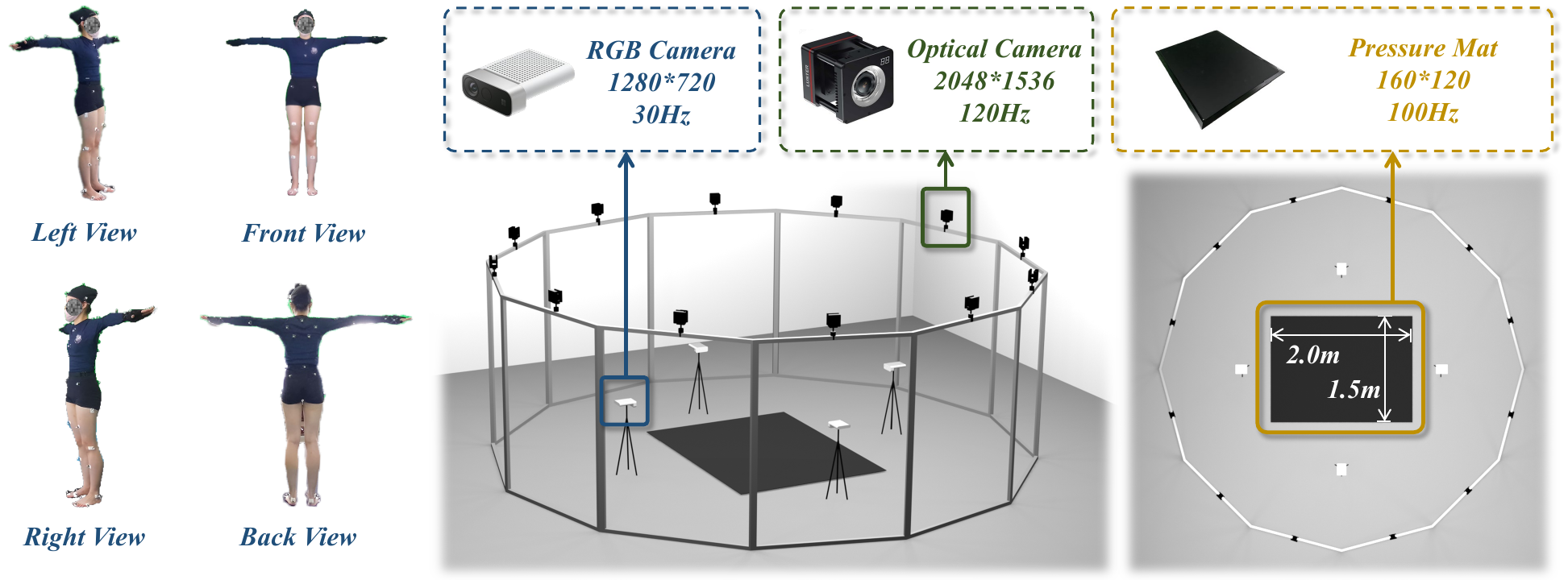}
\caption{The architecture of our motion capture system for dataset collection.}
\label{fig:system}
\end{figure}

To systematically investigate the role of pressure signals from human mocap to humanoid imitation, we construct a large-scale multimodal dataset, \textbf{MotionPRO}. As shown in Fig.~\ref{fig:system}, the dataset jointly captures synchronized RGB videos, pressure measurements, and high-precision optical motion capture data. Although the RGB cameras, pressure mat, and optical system operate at native frame rates of 30 Hz, 100 Hz, and 120 Hz, respectively, all modalities are temporally aligned and uniformly resampled to 30 Hz for training and evaluation, ensuring frame-level correspondence across the full pipeline.

\begin{figure}[t]
\centering
\includegraphics[width=0.9\linewidth]{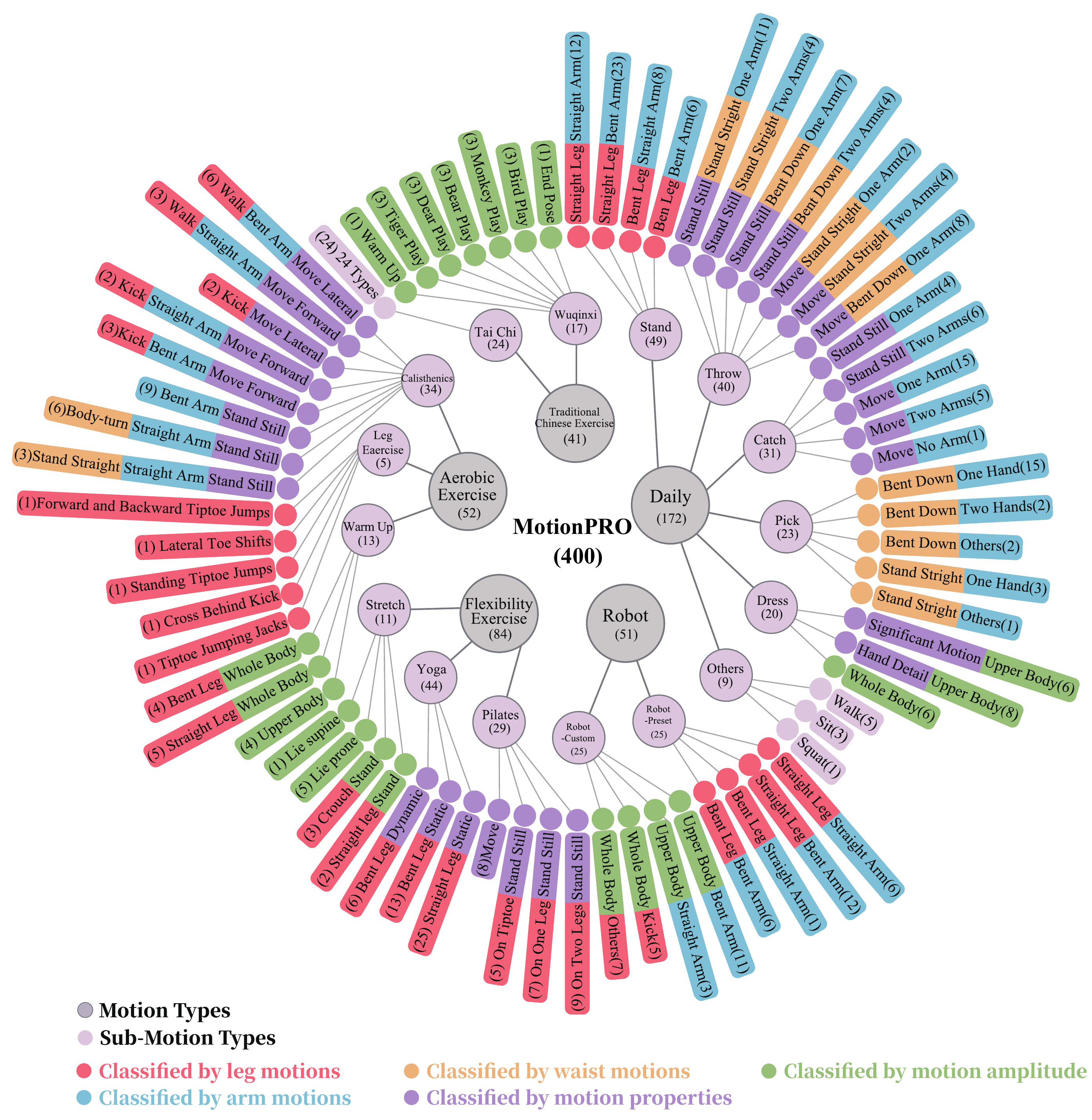}
\caption{Hierarchal distribution of 400 motion types.}
\label{fig:motion_type}
\end{figure}

MotionPRO contains 729 motion sequences with a total of 12.4M frames collected from 70 subjects, and encompasses 400 motion categories spanning daily activities, traditional exercises, aerobic motions, flexibility training, and robot-oriented movements. As illustrated in Fig.~\ref{fig:motion_type}, the motion types are hierarchically organized, exhibiting substantial diversity and complexity, which facilitates robust generalization.

We adopt the SMPL model~\cite{loper2015smpl} obtained by fitting motion capture marker data with Mosh++~\cite{mahmood2019amass} as the unified human representation. 
In addition, we provide per-joint contact labels derived from pressure signals and geometric constraints. To determine whether joint \( j \in \boldsymbol{J} \) is in contact with the pressure mat, we vertically project it onto the ground and calculate the sum of pressure in the vicinity as $  P_{j}$, as well as the distance to the ground plane $D_j$. We annotate the contact label $C_j$ by using the following strategy:

\begin{equation}
C_{j} = 
\left\{  
\begin{alignedat}{2}
1 \quad &\text{if} P_{j} \ge \tau_1 \ \text{and} \ D_j \le \tau_2\\
0 \quad &\text{otherwise,}
\end{alignedat}  
\right. 
\end{equation}
where \(\tau_1\) and \(\tau_2\) are thresholds for each variables.

\subsection{Human Pose and Trajectory Estimation}
\subsubsection{Metrics}
To evaluate our method, we adopt the following metrics: Mean Per Joint Position Error (\textbf{MPJPE}), Procrustes-aligned Mean Per Joint Position Error (\textbf{PMPJPE}), Per Vertex Error (\textbf{PVE}), and Acceleration (\textbf{Accel}).

For comparison with RGB-based methods in terms of global trajectory, we follow~\cite{shin2024wham} and split sequences into 100-frame segments, aligning each segment with ground truth using either the first two frames (\textbf{WMPJPE}) or all frames (\textbf{WAMPJPE}). We additionally report globally rigid-aligned Root Translation Error (\textbf{RTE}) over the entire trajectory, motion \textbf{Jitter}, and Whole Body Contact Error (\textbf{WBCE}), which measures the average absolute height distance between the ground plane and the joints in contact.

For the ablation study, we report Global Trajectory error of root (\textbf{GTraj}), Global Mean Per Joint Position Error (\textbf{GMPJPE}), Lower body PMPJPE (\textbf{LPM}), and Foot Sliding (\textbf{FS}) during contact phases. The unit of Jitter is $dm/s^3$. The unit of Accel is $m/frame^2$. All other metrics are in $mm$.

\subsubsection{Comparison with Other RGB-based Methods}
We evaluate our method on the MotionPRO dataset and compare it with representative RGB-based motion capture approaches. The results are summarized in Tab.~\ref{tab:frappe_pose} and Tab.~\ref{tab:trans_evaluation}.

Overall, our method consistently outperforms existing approaches across pose estimation, trajectory recovery, and motion stability. As shown in Tab.~\ref{tab:frappe_pose}, our full model achieves the best performance on all pose-related metrics, demonstrating the effectiveness of incorporating pressure signals as complementary information to visual inputs.

Compared with purely vision-based methods, our approach shows clear advantages in global motion estimation. As evidenced in Tab.~\ref{tab:trans_evaluation}, our method achieves significantly lower trajectory errors and more stable predictions, indicating that pressure signals provide strong cues for modeling human--ground interactions, which are difficult to infer from RGB data alone. This leads to reduced drift and more accurate long-term motion reconstruction.

In addition, our method produces smoother motion sequences with improved physical plausibility, as reflected by lower Accel, Jitter and WBCE in Tab.~\ref{tab:frappe_pose} and Tab.~\ref{tab:trans_evaluation}. Such improvements are crucial for downstream applications such as humanoid motion imitation, where stability and physical plausibility are essential.

Notably, even the basic pressure fusion variant (FRAPPE) already outperforms most RGB-based baselines, while the full model ($\text{FRAPPE}^{++}$) further improves performance across all aspects. This demonstrates that pressure information serves as a strong physical prior and can be more effectively leveraged through proper feature representation and temporal modeling.

\begin{table}[tbp]
  \centering
  \resizebox{1\linewidth}{!}{
  \begin{tabular}{@{}lccccc@{}}
    \toprule
    Methods & MPJPE $\downarrow$ & PMPJPE $\downarrow$ & PVE $\downarrow$  & Accel $\downarrow$  \\
    \midrule
    VIBE~\cite{kocabas2020vibe} & 59.7 & 40.9 & 82.9    & 19.6  \\
    CLIFF~\cite{li2022cliff} & 54.7 & 39.7 &  68.6  & 24.3 \\
    SMPLer-X~\cite{cai2023smpler} & {51.6} & {32.8}  & 72.4  & 437.3  \\
    TRACE~\cite{sun2023trace} & 61.4 & 43.2 &  81.4  & 14.6 \\
    PhysPT~\cite{zhang2024physpt} & 56.4 & 38.7  & 72.6  & {3.0}  \\
    WHAM~\cite{shin2024wham} & 160.4 & \cellcolor{Second}{28.3}  & 227.5  & 2.9  \\
    GVHMR~\cite{shen2024gvhmr} & 65.0 & 43.5  & 71.5  & \cellcolor{Second}1.6  \\
    FRAPPE~\cite{DBLP:conf/cvpr/RenLHZZYSC25}  & \cellcolor{Second}{41.8} & 30.2 & \cellcolor{Second}58.6    & {3.0}  \\
    \midrule
    $\text{FRAPPE}^{++}$  & \cellcolor{First}{33.2} & \cellcolor{First}24.7 & \cellcolor{First}47.5    & \cellcolor{First}{1.5}  \\    
    \bottomrule
  \end{tabular}
  }
  \vspace{1pt}
  \caption{Evaluation of global pose estimation on MotionPRO.}
  \label{tab:frappe_pose}
\vspace{-1.5em}
\end{table}

\begin{table}[tbp]
  \centering
  \resizebox{1\linewidth}{!}{
  \begin{tabular}{@{}lccccc@{}}
    \toprule
    Methods & WMPJPE $\downarrow$ & WAMPJPE $\downarrow$ & RTE $\downarrow$ & Jitter  $\downarrow$ & WBCE $\downarrow$  \\
    \midrule
    TRACE & 141.2 & 92.5 & 1193.0 & 68.6   &10272.4 \\

    WHAM & 75.6 & {50.2}  & 1023.0 & 9.2  & 1217.6  \\
    GVHMR & 75.0 & 57.1 & 507.1 & \cellcolor{First}{1.8} & 458.3  \\  FRAPPE  & \cellcolor{Second}{60.8} & \cellcolor{Second}44.6 & \cellcolor{Second}41.6 & 6.0  & \cellcolor{Second}{110.2}  \\   
    \midrule
    $\text{FRAPPE}^{++}$  & \cellcolor{First}{45.8} & \cellcolor{First}32.4 & \cellcolor{First}25.9 & \cellcolor{Second}2.2  & \cellcolor{First}{98.8}  \\
    
    \bottomrule
  \end{tabular}
  }
  \vspace{1pt}
  \caption{Evaluation of global trajectory on MotionPRO. }
  \vspace{-0.5em}
  \label{tab:trans_evaluation}

\end{table}

\begin{figure}[t]
\centering
\includegraphics[width=\linewidth]{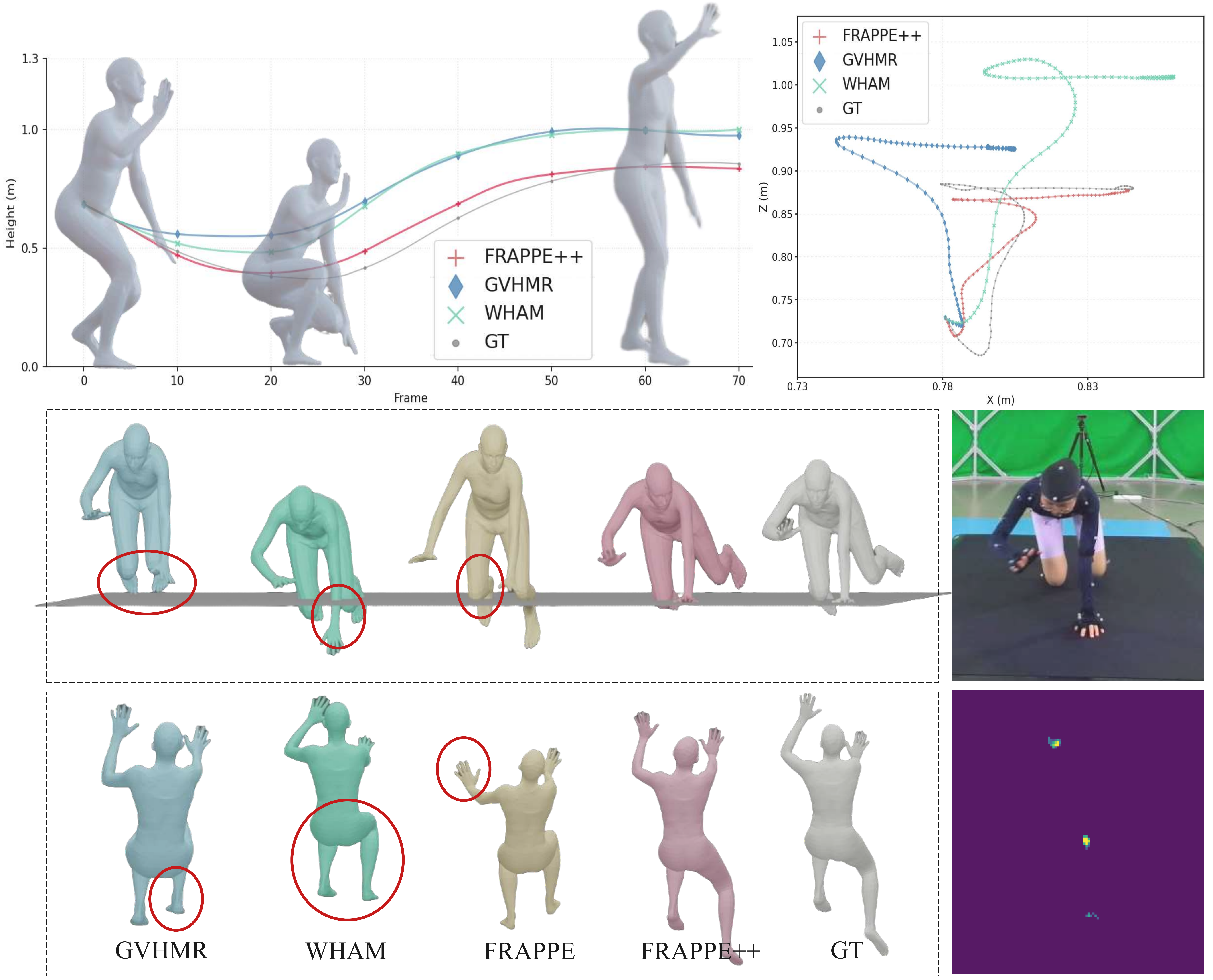}  
\caption{Qualitative comparison with methods for human pose estimation.}  
\label{fig:wpp}   
\vspace{-1em}
\end{figure}

\begin{table*}[tbp]
\centering
\resizebox{\textwidth}{!}{
\begin{tabular}{lccccccccc}
\hline
Models & MPJPE $\downarrow$ & PMPJPE $\downarrow$ & LPM. & GTraj $\downarrow$ & GMPJPE $\downarrow$ & Jitter $\downarrow$ $\downarrow$ & Accel $\downarrow$ & FS $\downarrow$ \\
\hline
w/o Pressure & 65.3 & 49.4 & 46.2 & 209.6 & 213.7 & 4.3 & 2.3 & 5.4 \\
w/o SPE & \cellcolor{First}31.9 & \cellcolor{First}24.0 & \cellcolor{First}23.5 & \cellcolor{Second}64.0 & \cellcolor{First}64.0 & \cellcolor{Second}2.4 & \cellcolor{First}1.5 & \cellcolor{First}2.1 \\
w/o TCAM & 39.0 & 27.1 & 26.8 & 67.5 & 70.6 & 6.3 & 2.4 & 3.3 \\
Full Model (Ours) & \cellcolor{Second}33.2 & \cellcolor{Second}24.7 & \cellcolor{Second}24.4 & \cellcolor{First}61.3 & \cellcolor{First}64.0 & \cellcolor{First}2.2 & \cellcolor{First}1.5 & \cellcolor{First}2.1 \\
\hline
\end{tabular}
}
\vspace{1pt}
\caption{Ablation study on MotionPRO.}
\label{tab:motionpro_ablation}
\end{table*}

\subsubsection{Ablation Study}

We conduct ablation studies on MotionPRO to analyze the contribution of each key component. The results are shown in Tab.~\ref{tab:motionpro_ablation}.

Removing the pressure modality leads to a significant performance drop across all metrics. The degradation is particularly pronounced in trajectory estimation and lower-body pose accuracy, indicating that pressure signals provide essential information about contact and support. This is especially important for resolving foot--ground interactions, which are difficult to infer from visual inputs alone.

Temporal modeling also plays a critical role in motion estimation. Removing TCAM leads to degradation in both pose and trajectory estimation accuracy, along with noticeable increases in jitter and acceleration errors. This indicates that modeling temporal dependencies not only improves motion estimation performance , but is also essential for maintaining motion smoothness and consistency.

Pressure encoding mainly affects trajectory estimation. When SPE is removed, trajectory-related metrics degrade noticeably, while pose accuracy shows a slight improvement. This suggests a trade-off between learning strong pressure representations and optimizing pose estimation.

\subsection{Pressure-Guided Humanoid Motion Imitation}
\subsubsection{Experiment Setup}
We deploy and evaluate our method on the Unitree G1 humanoid robot. To ensure a fair and consistent evaluation, all control policies are trained with the same number of parallel environments (4,096) and transferred to MuJoCo for unified assessment, as well as to the real Unitree G1 platform without any fine-tuning, enabling cross-simulator and sim-to-real transferability analysis. Each baseline is trained following its original recommended configuration until convergence.

\subsubsection{Metrics}
To evaluate the performance of humanoid motion imitation, we adopt the following metrics.

Success Rate (\textbf{SR}) measures the proportion of successful trials across all motion sequences, where each sequence is executed for 20 independent trials. A trial is considered successful if the robot does not fall during execution and the positional deviation between its root joint and the reference trajectory remains below 0.85\,m throughout. 
Motion Completion (\textbf{MC}) computes the ratio of successfully completed frames to the total number of reference frames. These two metrics jointly assess the stability of the control policy.

Foot Contact Accuracy (\textbf{FCA}) measures the percentage of frames in which the robot's foot contact states match the reference contact labels derived from pressure measurements recorded during human demonstration.

To evaluate end-to-end trajectory-level accuracy, we use Root Position Error (\textbf{RPE}, $m$) and Root Velocity Error (\textbf{RVE}, $m/s$) to quantify the mean Euclidean distance and velocity difference at the robot's root joint during execution, respectively, where RPE is computed after aligning the executed and reference trajectories at the first frame to remove initial positional offset. 
World-frame Mean Per-Joint Position Error (\textbf{WMPJPE}, $m$) measures joint-level positional deviations in the world coordinate frame, similarly computed after first-frame alignment to remove initial positional offset.

To evaluate the similarity between the robot's local pose and that of the human demonstrator, we adopt Mean Joint Angle Error (\textbf{MJAE}, $rad$) to measure joint angular deviation, and Procrustes-aligned Mean Per-Joint Position Error (\textbf{PMPJPE}, $m$) to evaluate local joint position similarity after removing global translational, rotational, and scale discrepancies via Procrustes alignment.

Each metric is averaged across all 20 trials per motion sequence. 
Since methods with lower MC only execute a fraction of the motion sequence, their error metrics are computed over fewer and typically easier frames, leading to artificially low values. To correct for this bias, all error metrics (RPE, RVE, WMPJPE, MJAE, and PMPJPE) are normalized by MC before comparison.

\begin{table*}[thbp]
\centering
\resizebox{\textwidth}{!}{
\begin{tabular}{llcccccccc}
\toprule
Reference & Control Policy & SR $\uparrow$ & MC $\uparrow$ & FCA $\uparrow$ & RPE $\downarrow$ & RVE $\downarrow$ & WMPJPE $\downarrow$ & MJAE $\downarrow$ & PMPJPE $\downarrow$ \\
\midrule
Optical MoCap & BeyondMimic & \cellcolor{First}0.9278 & \cellcolor{Second}0.9727 & \cellcolor{First}0.7008 & \cellcolor{Second}0.2706 & \cellcolor{Second}0.2594 & \cellcolor{Second}0.5092 & \cellcolor{Second}0.5533 & \cellcolor{First}0.0830 \\
WHAM               & BeyondMimic & 0.0333 & 0.0616 & 0.0244 & 1.5763 & 9.1591 & 6.9643 & 3.3961 & 1.7045 \\
GVHMR              & BeyondMimic & 0.2278 & 0.4644 & 0.2885 & 0.5286 & 0.8417 & 1.0913 & 0.9464 & 0.1996 \\
$\text{FRAPPE}^{++}$ (Ours) & BeyondMimic & \cellcolor{Second}0.9056 & \cellcolor{First}0.9746 & \cellcolor{Second}0.6883 & \cellcolor{First}0.2150 & \cellcolor{First}0.2295 & \cellcolor{First}0.4796 & \cellcolor{First}0.4766 & \cellcolor{Second}0.0875 \\
\bottomrule
\end{tabular}
}
\vspace{1pt}
\caption{Quantitative Evaluation of the Impact of Human MoCap Quality on Motion Imitation Performance.}
\label{tab:humanoid_mujoco_reference}
\end{table*}

\begin{table*}[thbp]
\centering
\resizebox{\textwidth}{!}{
\begin{tabular}{llcccccccc}
\toprule
Reference & Control Policy & SR $\uparrow$ & MC $\uparrow$ & FCA $\uparrow$ & RPE $\downarrow$ & RVE $\downarrow$ & WMPJPE $\downarrow$ & MJAE $\downarrow$ & PMPJPE $\downarrow$ \\
\midrule
Optical MoCap & PBHC & 0.0381 & 0.1710 & 0.0829 & 1.0474 & 3.300 & 1.1561 & 1.7099 & 0.7000 \\
Optical MoCap & BeyondMimic & \cellcolor{First}0.9278 & \cellcolor{First}0.9727 & \cellcolor{Second}0.7008 & \cellcolor{Second}0.2706 & \cellcolor{Second}0.2594 & \cellcolor{Second}0.5092 & \cellcolor{First}0.5533 & \cellcolor{First}0.0830 \\
Optical MoCap & PSP (Ours) & \cellcolor{Second}0.9222 & \cellcolor{Second}0.9725 & \cellcolor{First}0.7262 & \cellcolor{First}0.2592 & \cellcolor{First}0.2486 & \cellcolor{First}0.4950 & \cellcolor{Second}0.5534 & \cellcolor{Second}0.0854 \\
\midrule
GVHMR              & PBHC & 0.0452 & 0.1727 & 0.0880 & 1.0301 & 3.2131 & 1.1210 & 1.7336 & 0.7041 \\
GVHMR              & BeyondMimic & \cellcolor{Second}0.2278 & \cellcolor{Second}0.4644 & \cellcolor{Second}0.2885 & \cellcolor{Second}0.5286 & \cellcolor{Second}0.8417 & \cellcolor{Second}1.0913 & \cellcolor{Second}0.9464 & \cellcolor{Second}0.1996 \\
GVHMR              & PSP (Ours) & \cellcolor{First}0.5778 & \cellcolor{First}0.7142 & \cellcolor{First}0.4726 & \cellcolor{First}0.3988 & \cellcolor{First}0.5000 & \cellcolor{First}0.7415 & \cellcolor{First}0.6606 & \cellcolor{First}0.1395 \\
\midrule
$\text{FRAPPE}^{++}$ (Ours) & PBHC & 0.0667 & 0.1644 & 0.0940 & 1.1107 & 3.3978 & 1.1807 & 1.8327 & 0.7263 \\
$\text{FRAPPE}^{++}$ (Ours) & BeyondMimic & \cellcolor{Second}0.9056 & \cellcolor{Second}0.9746 & \cellcolor{Second}0.6883 & \cellcolor{Second}0.2150 & \cellcolor{Second}0.2295 & \cellcolor{Second}0.4796 & \cellcolor{Second}\cellcolor{Second}0.4766 & \cellcolor{Second}0.0875 \\
$\text{FRAPPE}^{++}$ (Ours) & PSP (Ours) & \cellcolor{First}0.9444 & \cellcolor{First}0.9846 & \cellcolor{First}0.6918 & \cellcolor{First}0.2120 & \cellcolor{First}0.2144 & \cellcolor{First}0.4737 & \cellcolor{First}0.4756 & \cellcolor{First}0.0873 \\
\bottomrule
\end{tabular}
}
\vspace{1pt}
\caption{Quantitative Evaluation of the Impact of Motion Control on Motion Imitation Performance.}
\label{tab:humanoid_mujoco_control}
\end{table*}

\subsubsection{Effect of Reference Motion Quality}
Tab.~\ref{tab:humanoid_mujoco_reference} presents quantitative results of humanoid motion imitation under different reference motion sources. We selected BeyondMimic~\cite{liao2025beyondmimic}, a baseline method with good control performance, as the control policy.

\begin{figure*}[htb]
	\centering
	\includegraphics[width=\linewidth]{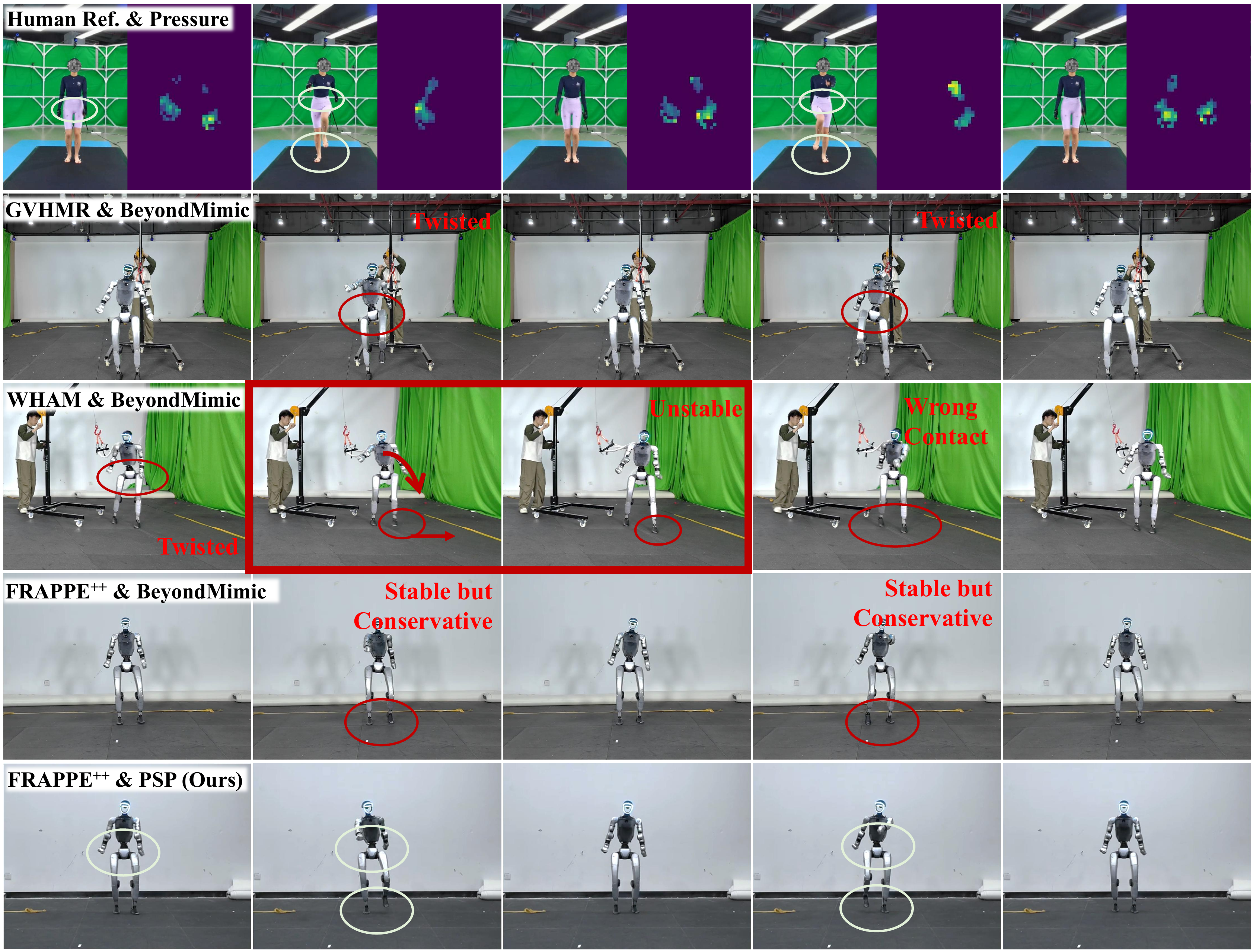}
	\caption{\textbf{Real-world humanoid motion imitation results.} Rows show execution results under different reference sources and control policies. Low-quality references (GVHMR, WHAM) lead to twisted postures and instability, while FRAPPE$^{++}$ with BeyondMimic achieves stable but conservative foot placement. FRAPPE$^{++}$ with PSP (Ours) produces stable execution with more natural foot contact dynamics (white circles), validating sim-to-real transfer of pressure supervision.}
	\label{fig:real_env}
\end{figure*}

We find that reference quality directly governs imitation performance. 
Optical MoCap and FRAPPE$^{++}$ achieve strong SR of 0.9278 and 0.9056, MC of 0.9727 and 0.9746, and FCA of 0.7008 and 0.6883 respectively, while GVHMR~\cite{shen2024gvhmr} degrades substantially (SR: 0.2278) and WHAM~\cite{shin2024wham} nearly collapses (SR: 0.0333). After MC normalization, WHAM exhibits severely inflated error metrics (RPE: 1.5763, RVE: 9.1591, WMPJPE: 6.9643), reflecting early termination rather than genuine motion execution. 
Notably, FRAPPE$^{++}$ achieves comparable performance to Optical MoCap across all metrics, demonstrating that our RGB and pressure fusion-based mocap method produces reference motions of sufficient quality to serve as a viable alternative to optical motion capture. As shown in Fig.~\ref{fig:real_env}, GVHMR and WHAM references introduce severe kinematic distortions and incorrect foot contact in real-world execution, leading to postural instability and even falls, whereas FRAPPE$^{++}$ enables the robot to maintain stable locomotion and accurate posture throughout execution.

The lower RPE, RVE, WMPJPE, and MJAE of FRAPPE$^{++}$ compared to Optical MoCap indicate that the control policy tracks the FRAPPE$^{++}$ reference more accurately. We attribute this to the inherently smoother kinematics of the FRAPPE$^{++}$ reference: the RGB and pressure fusion-based reconstruction attenuates high-frequency motion details, yielding reference trajectories that better conform to the robot's dynamic constraints and are therefore more amenable to precise tracking.

\subsubsection{Effect of Pressure-Supervised Control}
Tab.~\ref{tab:humanoid_mujoco_control} presents quantitative results under three control policies across all reference sources. As WHAM-based references consistently fail to produce stable locomotion across all control policies (SR $<$ 0.05), we exclude it from further analysis and focus on the remaining reference sources. PBHC~\cite{xie2025kungfubot} fails to produce stable locomotion regardless of reference quality, with SR below 0.07 in all conditions, indicating that a capable base policy is a prerequisite for effective motion imitation. PSP consistently outperforms BeyondMimic across all estimated reference sources (GVHMR and FRAPPE$^{++}$), demonstrating that pressure supervision provides effective stabilizing contact constraints throughout the motion imitation pipeline.

Under the FRAPPE$^{++}$ reference, PSP achieves the best overall performance: SR improves from 0.9056 to 0.9444 and MC from 0.9746 to 0.9846, marginally surpassing the Optical MoCap-based PSP counterpart (SR: 0.9222, MC: 0.9725). We attribute this to the unified pressure-guided design that runs through both the motion capture and imitation stages: the pressure signal shapes the FRAPPE$^{++}$ reference with physically grounded contact constraints during capture, and PSP reinforces the same contact dynamics during policy training, resulting in natural alignment between the two stages and leading to more complete and stable execution.

Under the Optical MoCap reference, PSP yields a slight decrease in SR and MC compared to BeyondMimic, while improving FCA and all error metrics. We attribute the marginal SR drop to a cross-modal misalignment between the optical motion capture system and the pressure sensing system. When both references are of exceptionally high fidelity, subtle spatial and temporal misalignments between the two modalities introduce conflicting supervision signals during policy training, occasionally destabilizing execution near support leg transitions. This phenomenon is unique to the Optical MoCap setting. Under FRAPPE$^{++}$, the reference inherently carries pressure-informed contact constraints through the fusion process, resulting in better natural alignment between the two modalities and allowing PSP to consistently improve performance.

Real-world results further corroborate these findings: as shown in Fig.~\ref{fig:real_env}, FRAPPE$^{++}$ with BeyondMimic achieves stable execution but exhibits conservative foot placement that deviates from the reference contact pattern, whereas FRAPPE$^{++}$ with PSP produces more natural foot contact dynamics and weight transfer, as highlighted by the white circles.

\section{Conclusion}

In this paper, we investigate the role of pressure as a physical grounding signal for humanoid motion imitation and presented PressMimic, a unified framework that integrates pressure into both motion perception and control. By augmenting vision-based motion capture with pressure, our approach improves the accuracy and physical plausibility of whole-body human motion estimation. Furthermore, by incorporating pressure-derived supervision into policy learning, we enable humanoid robots to better reproduce contact dynamics, leading to more stable and reliable execution.

Our study highlights that pressure provides complementary information to visual inputs by explicitly encoding human–environment interactions, which are otherwise difficult to infer from images alone. This additional modality not only reduces ambiguity in motion estimation but also serves as an effective supervisory signal for control, bridging the long-standing gap between kinematic imitation and physically consistent execution.

Despite these advances, several limitations remain. First, the pressure signals are currently limited to ground interactions and may not fully capture complex multi-contact scenarios involving interactions with external objects. Second, the integration of pressure in robot control is still mainly applied to bipedal contact-related scenarios, leaving room for more rigorous whole-body contact dynamics modeling, mapping, and learning. Future work will explore richer physical representations, improved generalization beyond instrumented environments, and tighter coupling between perception and control. 
We believe that incorporating physical sensing modalities such as pressure is a crucial step toward achieving robust and physically grounded embodied intelligence.

\section*{Acknowledgments}
This work was supported in part by the National
Natural Science Foundation of China under Grant U25B2046 and 62231002.



 
%

\bibliographystyle{IEEEtran}
\bibliography{IEEEexample}


 

\vspace{-33pt}
\begin{IEEEbiography}[{\includegraphics[width=1in,height=1.25in,clip,keepaspectratio]{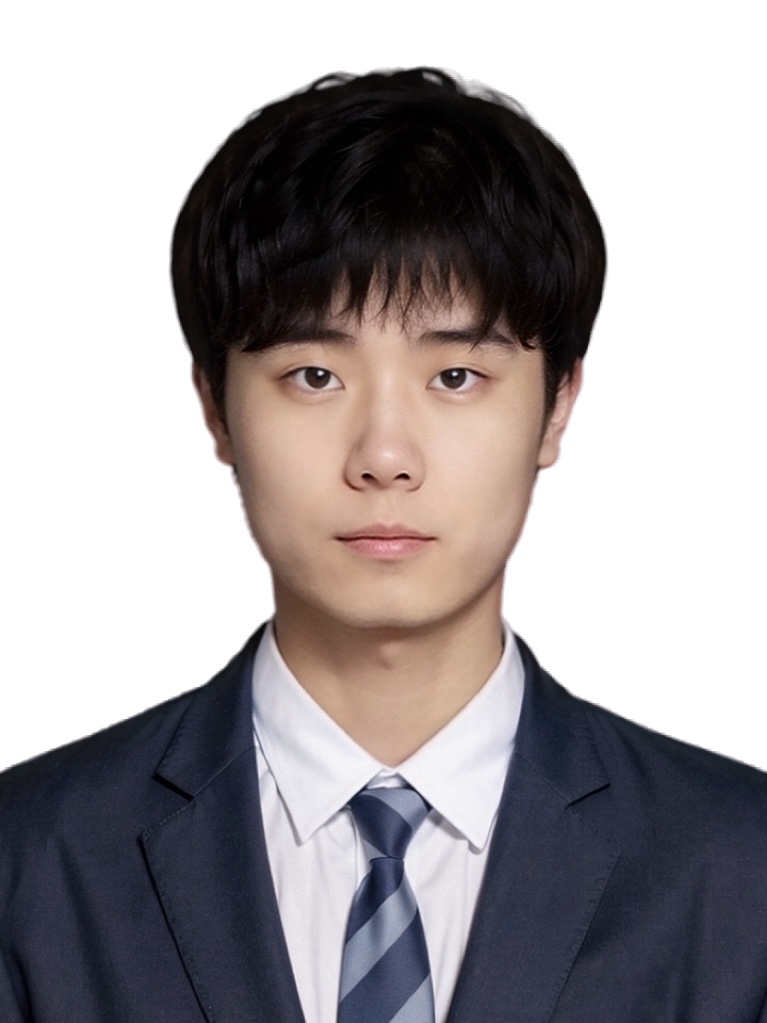}}]{Yi Lu}
is a graduate student for Ph.D degree at the School of Electronic Science and Engineering, Nanjing University, China. He received his B.S. degree from Nanjing University of Posts and Telecommunications in 2021. His research interests include embodied intelligence, humanoid robot motion control, and human motion capture.
\end{IEEEbiography}
\vspace{-33pt}
\begin{IEEEbiography}
[{\includegraphics[width=1in,height=1.25in,clip,keepaspectratio]{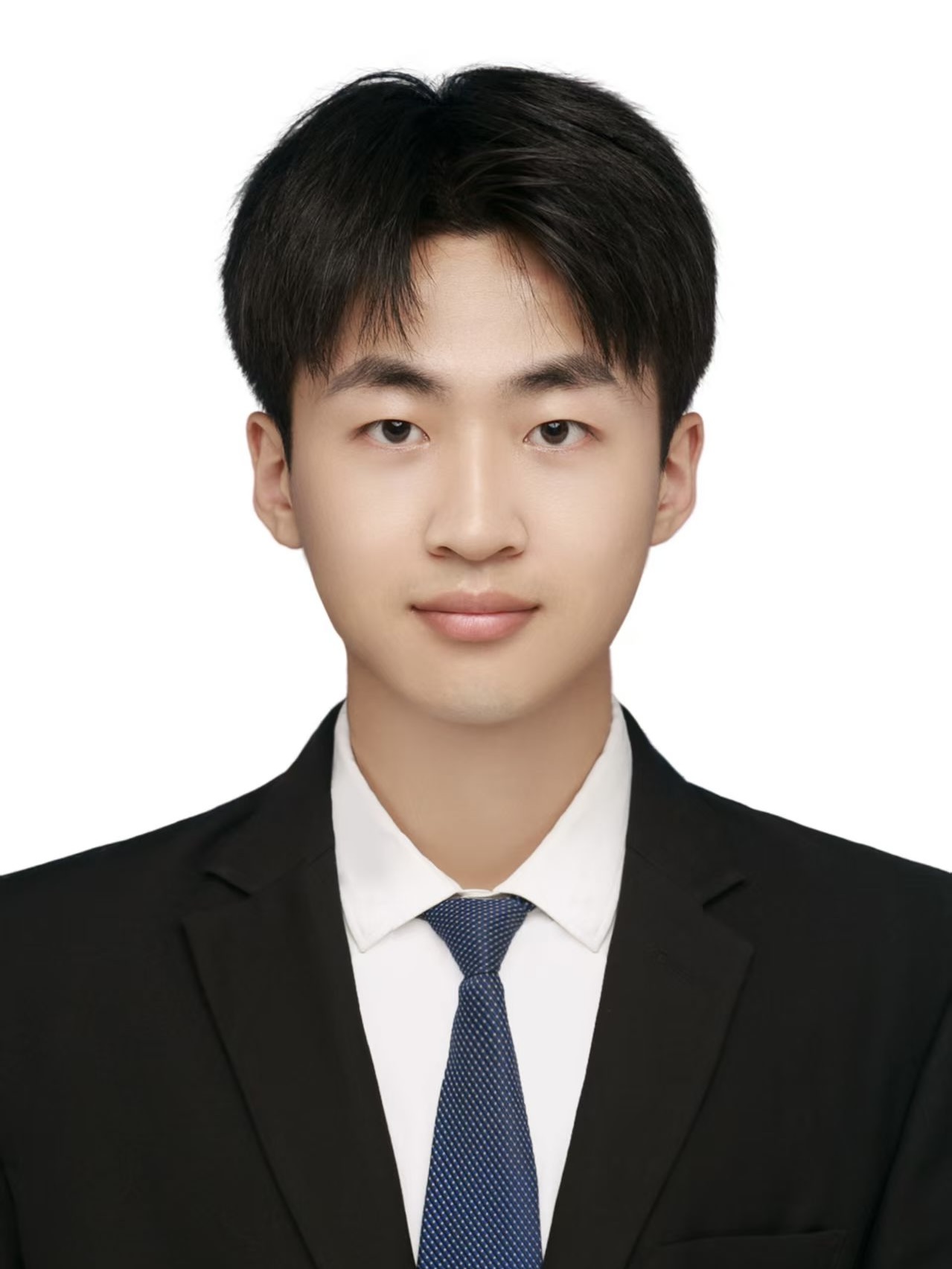}}]{Shenghao Ren}
is a postgraduate student at the School of Electronic Science and Engineering, Nanjing University, China. He received a B.S. degree from Nanjing University in 2023. His research interests include computer vision and human mesh recovery.
\end{IEEEbiography}
\vspace{-33pt}
\begin{IEEEbiography}
[{\includegraphics[width=1in,height=1.25in,clip,keepaspectratio]{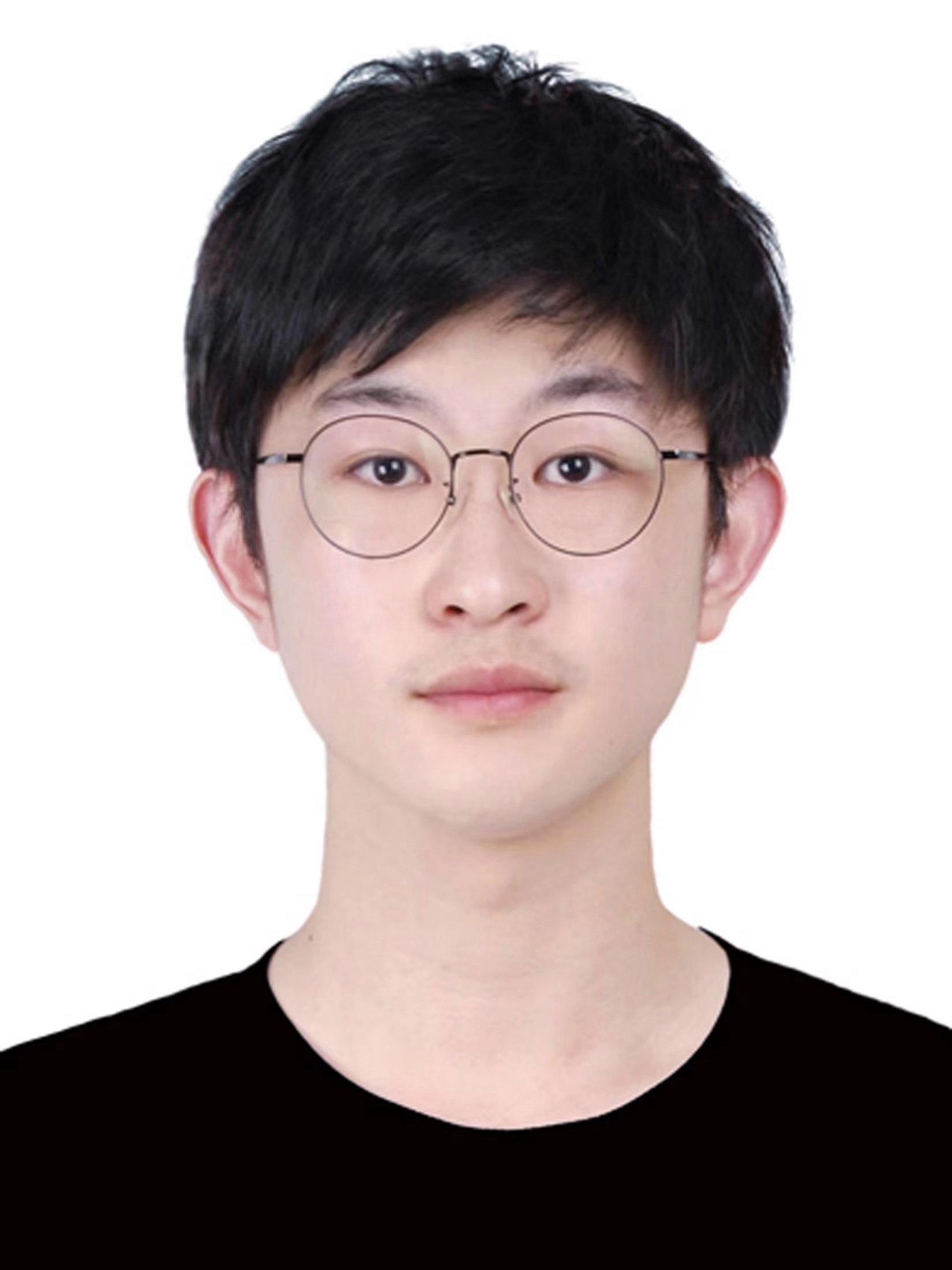}}]{Tianyu Xiong}
is an undergraduate student at the School of Electronic Science and Engineering, Nanjing University, China. His research interests include embodied intelligence and human motion capture.
\end{IEEEbiography}
\vspace{-33pt}
\begin{IEEEbiography}
[{\includegraphics[width=1in,height=1.25in,clip,keepaspectratio]{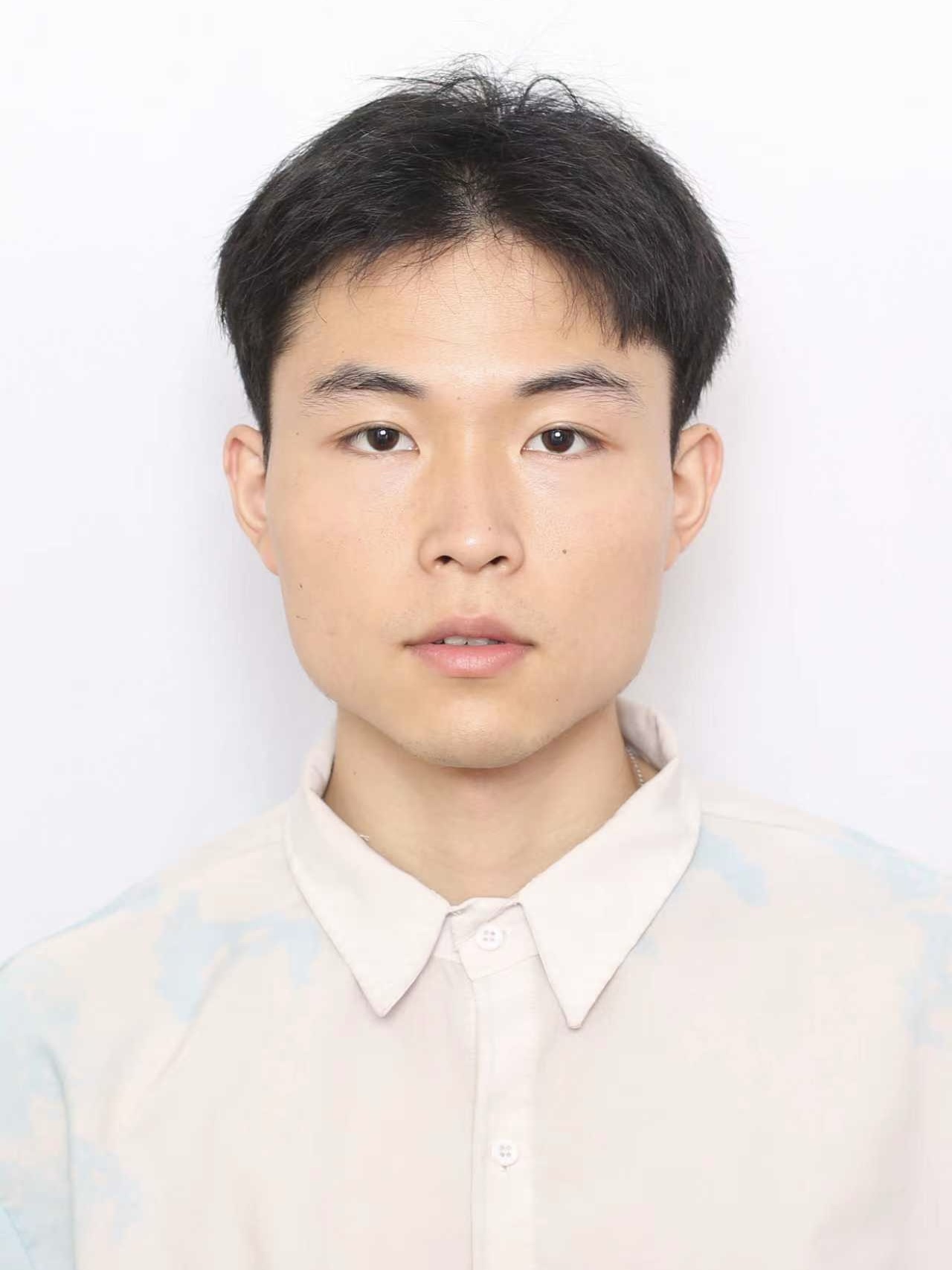}}]{Zhaoxiang Li}
is an undergraduate student at the School of Electronic Science and Engineering, Nanjing University, China. His research interests include reinforcement learning and embodied intelligence.
\end{IEEEbiography}
\vspace{-33pt}
\begin{IEEEbiography}
[{\includegraphics[width=1in,height=1.25in,clip,keepaspectratio]{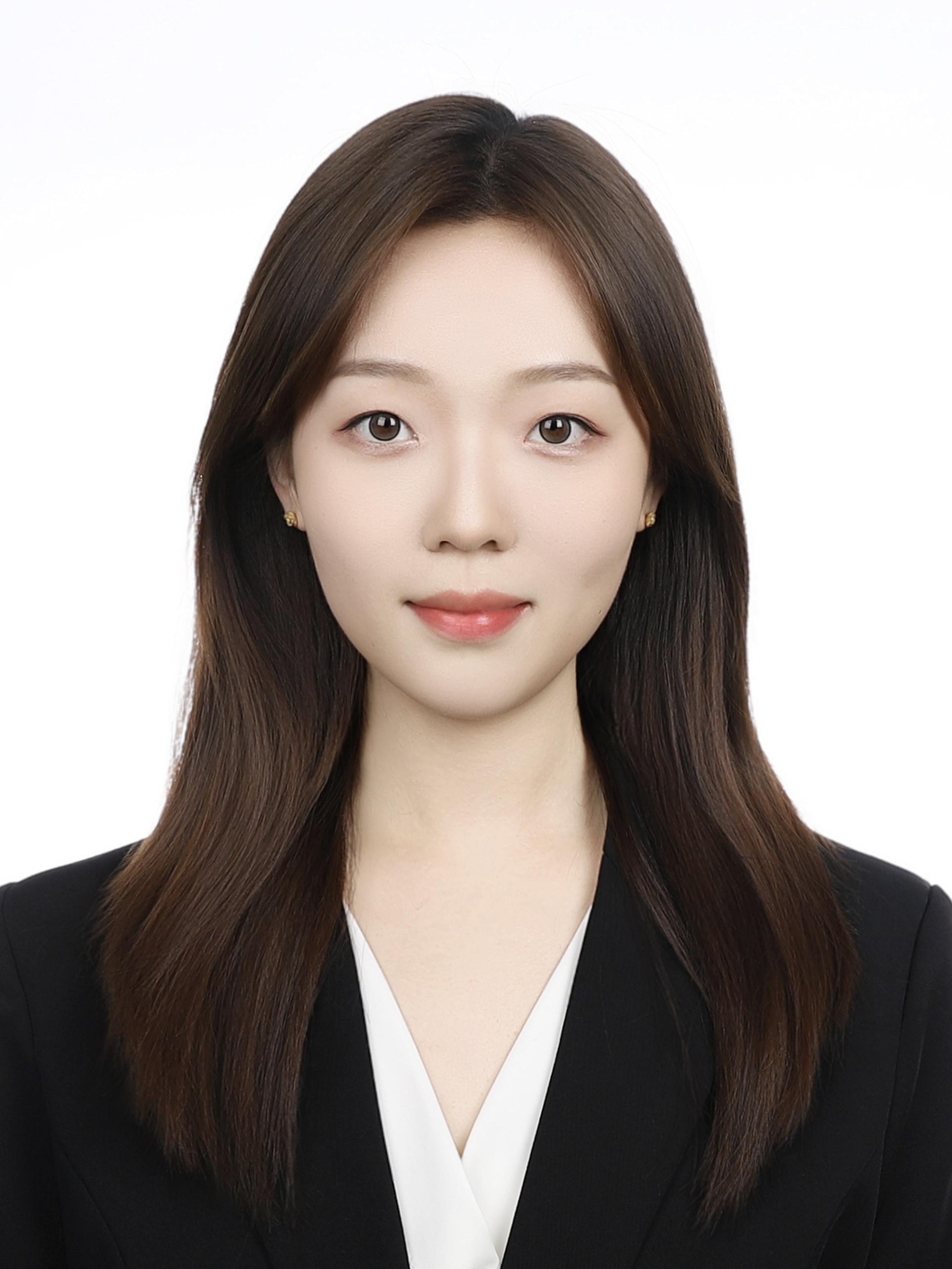}}]{Jiaqi Li}
is an undergraduate student at the School of Electronic Science and Engineering, Nanjing University, China.Her research interests include humanoid robot motion control and evaluation.
\end{IEEEbiography}
\vspace{-33pt}
\begin{IEEEbiography}
[{\includegraphics[width=1in,height=1.25in,clip,keepaspectratio]{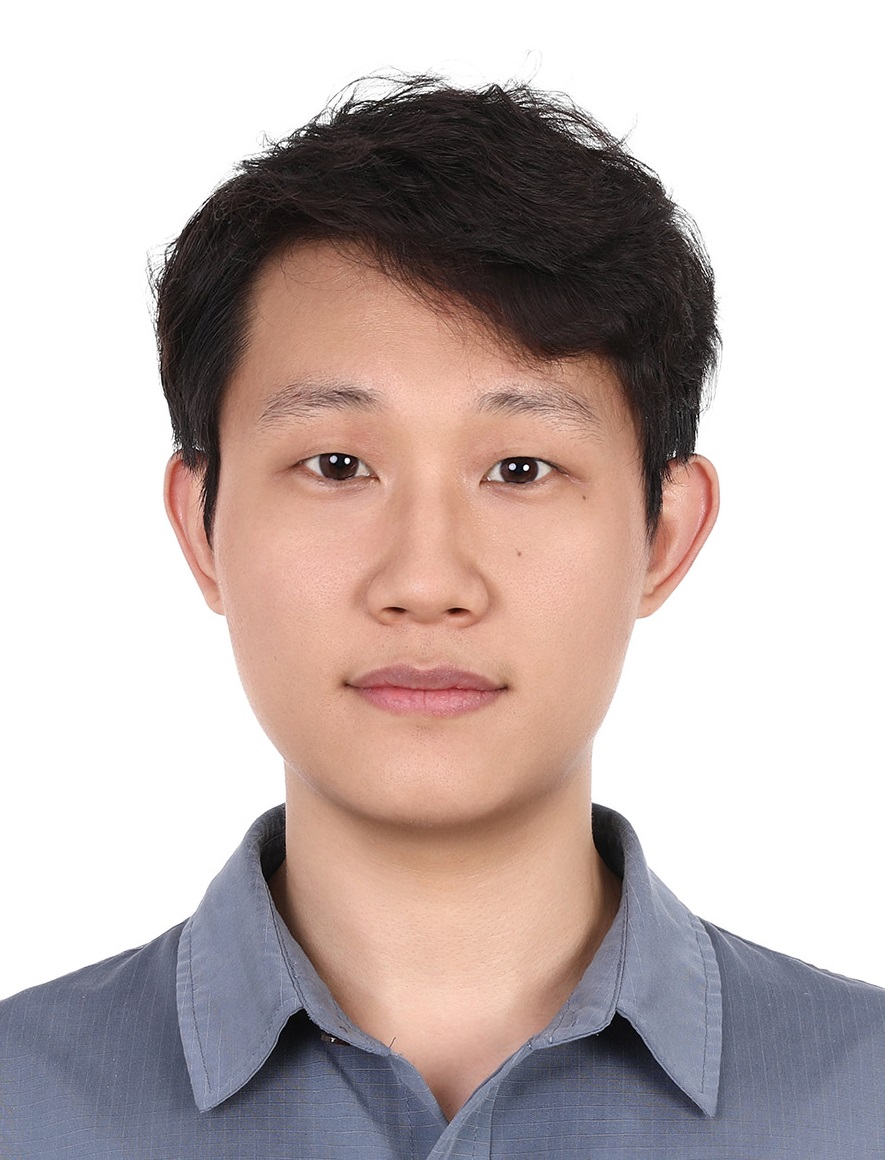}}]{He Zhang}
received the M.S. and Ph.D. degrees from Beihang University in 2018 and 2024, respectively. He is currently a postdoc at BNRist, Tsinghua University, Beijing, China. His current research focuses on motion capture, biomechanics, and behaviour science.
\end{IEEEbiography}
\vspace{-30pt}
\begin{IEEEbiography}
[{\includegraphics[width=1in,height=1.25in,clip,keepaspectratio]{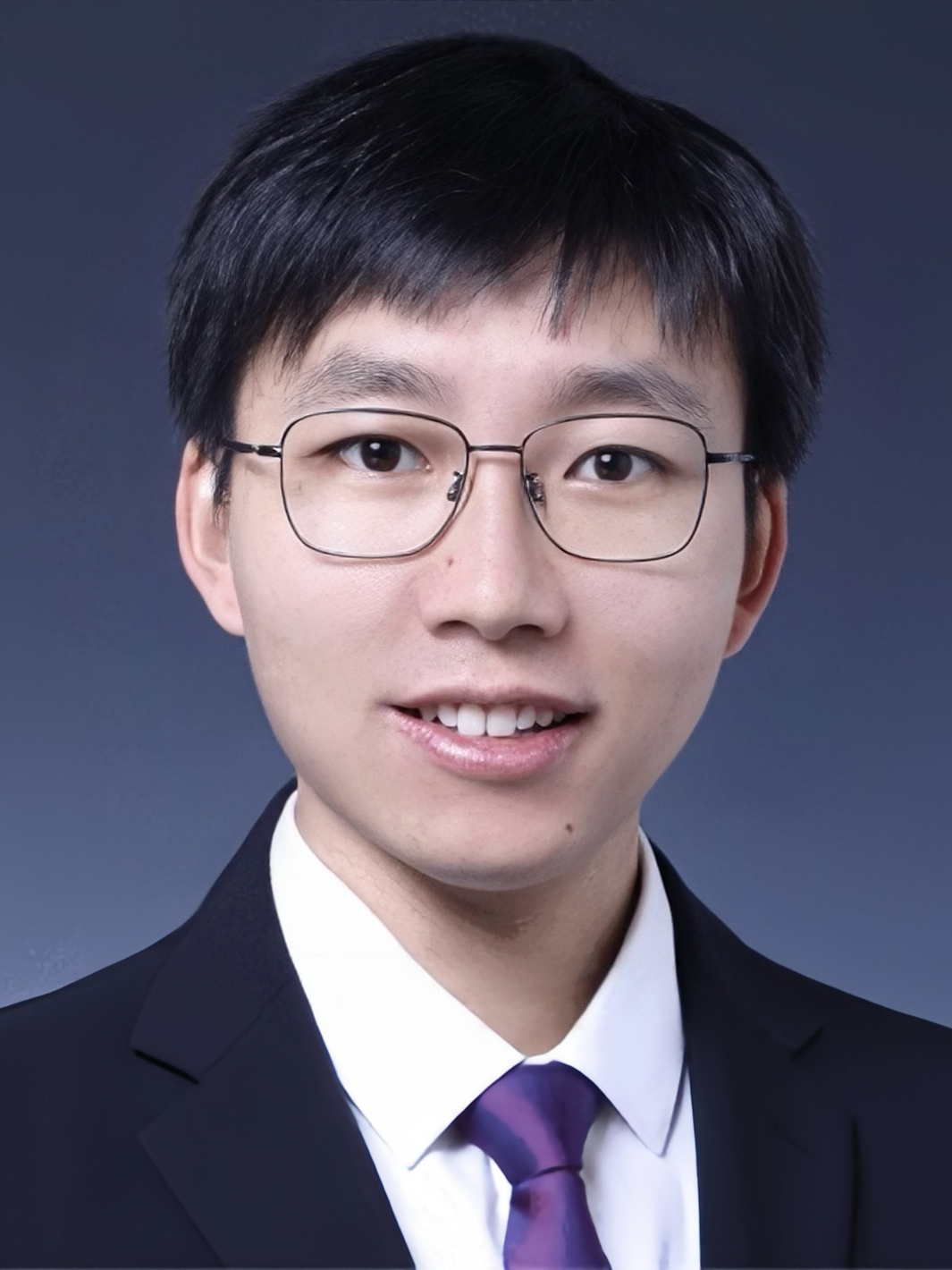}}]{Tao Yu}
is an associate researcher at BNRist, Tsinghua University. He received the B.S. degree in measurement and control from the Hefei University of Technology, in 2012, and the Ph.D degree from Beihang University, in 2019. His current research interests include 3D vision, AI, and computer graphics.
\end{IEEEbiography}
\vspace{-30pt}
\begin{IEEEbiography}
[{\includegraphics[width=1in,height=1.25in,clip,keepaspectratio]{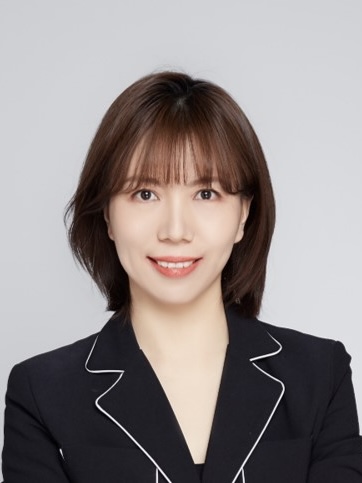}}]{Qiu Shen}
received the B.S. and Ph.D. degree from the University of Science and Technology of China, in 2004 and 2009 respectively. From 2009 to 2016, he has been with Huawei 2012 Lab and Nanjing University of Aeronautics and Astronautics. She is an associate professor at the School of Electronic Science and Engineering School, Nanjing University, China. Her current research focuses on next-generation video coding, collaborative video compression and analysis, embodied intelligence, and vision model for virtual reality.
\end{IEEEbiography}
\vspace{-30pt}
\begin{IEEEbiography}
[{\includegraphics[width=1in,height=1.25in,clip,keepaspectratio]{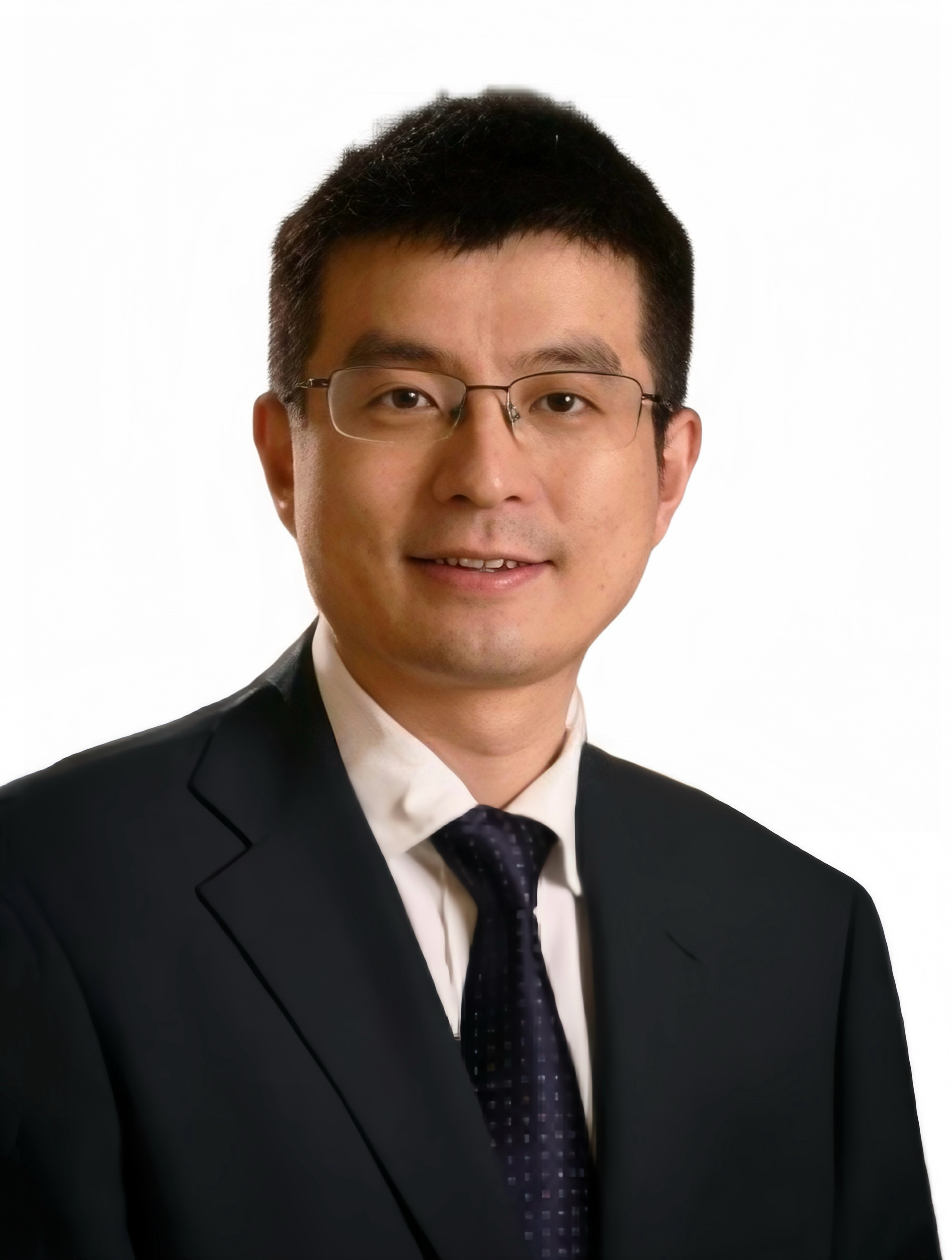}}]{Xun Cao}
received the B.S. degree from Nanjing University, Nanjing, China, in 2006, and the Ph.D. degree from the Department of Automation, Tsinghua University, Beijing, China, in 2012. He held visiting positions with Philips Research, Aachen, Germany, in 2008 and Microsoft Research Asia, Beijing, from 2009 to 2010. He was a Visiting Scholar with the University of Texas at Austin, Austin, TX, USA, from 2010 to 2011. He is a Professor at the School of Electronic Science and Engineering, Nanjing University. His current research interests include computational photography and image-based modeling and rendering.
\end{IEEEbiography}

\vspace{11pt}


\vfill

\end{document}